\documentclass[11pt, a4paper, logo, copyright]{googlecloud}

\pdftrailerid{redacted}

\makeatletter
\renewcommand\bibentry[1]{\nocite{#1}{\frenchspacing\@nameuse{BR@r@#1\@extra@b@citeb}}}
\makeatother

\usepackage{kantlipsum, lipsum}
\usepackage{dsfont}

\usepackage[authoryear, sort&compress, round]{natbib}

\usepackage[utf8]{inputenc} 
\usepackage[T1]{fontenc}    
\usepackage{url}            
\usepackage{booktabs}       
\usepackage{nicefrac}       
\usepackage{microtype}      
\usepackage{amsmath}
\usepackage{graphicx}
\usepackage{multicol}
\usepackage{hyperref}       
\usepackage[nameinlink]{cleveref}
\usepackage{bbm}
\usepackage{multirow}
\usepackage{soul}
\usepackage{float}
\usepackage{wrapfig}
\usepackage{blindtext}
\usepackage{tablefootnote}
\usepackage{amsfonts}
\usepackage[flushleft]{threeparttable}
\usepackage{bbding}
\usepackage{xcolor}
\usepackage{xspace}
\usepackage{bm}
\usepackage{arydshln}
\usepackage{subfigure}
\usepackage{enumitem}
\usepackage{setspace}
\usepackage{color}
\usepackage{algorithm}
\usepackage{longtable}

\usepackage[normalem]{ulem}
\usepackage{ulem}
\usepackage[nomargin,inline,marginclue,draft]{fixme}
\usepackage{balance}
\usepackage{verbatim}
\usepackage{diagbox}
\usepackage{changepage}
\usepackage{amssymb}
\usepackage{pifont}

\usepackage{amsmath,amsfonts,bm}

\def\eqref#1{equation~\ref{#1}}

\def\1{\bm{1}}

\DeclareMathAlphabet{\mathsfit}{\encodingdefault}{\sfdefault}{m}{sl}
\SetMathAlphabet{\mathsfit}{bold}{\encodingdefault}{\sfdefault}{bx}{n}

\usepackage{algpseudocode}

\usepackage{listings}
\usepackage{tcolorbox}
\tcbuselibrary{listings, breakable, skins}
\usepackage[table]{xcolor}

\newcommand{\lens}{\raisebox{-1pt}{\includegraphics[width=1.0em]{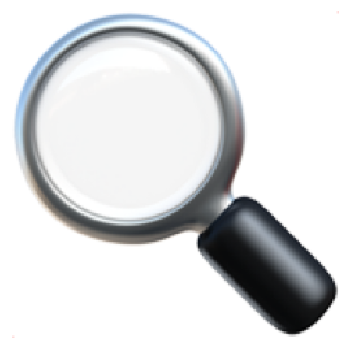}}}
\newcommand{\methodname}{\textsc{DocLens}\xspace}

\definecolor{promptbackground}{RGB}{235, 245, 255}
\definecolor{promptframe}{RGB}{60, 120, 180}
\definecolor{outputbackground}{gray}{0.95}
\definecolor{outputframe}{gray}{0.65}

\newtcblisting{promptbox}[2][]{
    listing only, breakable, title=#2,
    colback=promptbackground, colframe=promptframe, coltitle=white,
    fonttitle=\bfseries, boxrule=0.5pt, arc=3mm, colbacktitle=promptframe,
    fontupper=\small\ttfamily,
    #1
}

\title{\textsc{\methodname{}} \lens{}: A Tool-Augmented Multi-Agent Framework for Long Visual Document Understanding}

\correspondingauthor{lisujian@pku.edu.cn, jinsungyoon@google.com}

\author[1 2 *]{Dawei Zhu}
\author[2]{Rui Meng}
\author[2]{Jiefeng Chen}
\author[1]{Sujian Li}
\author[2]{Tomas Pfister}
\author[2]{Jinsung Yoon}

\affil[1]{School of Computer Science, Peking University}
\affil[2]{Google Cloud AI Research}

\newcommand{\projecturl}{https://dwzhu-pku.github.io/DocLens/}

\begin{abstract}

Comprehending long visual documents, where information is distributed across extensive pages of text and visual elements, is a critical but challenging task for modern Vision-Language Models (VLMs). Existing approaches falter on a fundamental challenge: evidence localization. They struggle to retrieve relevant pages and overlook fine-grained details within visual elements, leading to limited performance and model hallucination. To address this, we propose \methodname{}, a tool-augmented multi-agent framework that effectively ``zooms in'' on evidence like a lens. It first navigates from the full document to specific visual elements on relevant pages, then employs a sampling-adjudication mechanism to generate a single, reliable answer. Paired with Gemini-2.5-Pro, \methodname{} achieves state-of-the-art performance on MMLongBench-Doc and FinRAGBench-V, surpassing even human experts. The framework's superiority is particularly evident on vision-centric and unanswerable queries, demonstrating the power of its enhanced localization capabilities.

\end{abstract}

\begin{document}

\maketitle


\begin{figure}[h]
    \centering
    \includegraphics[width=1.0\textwidth]{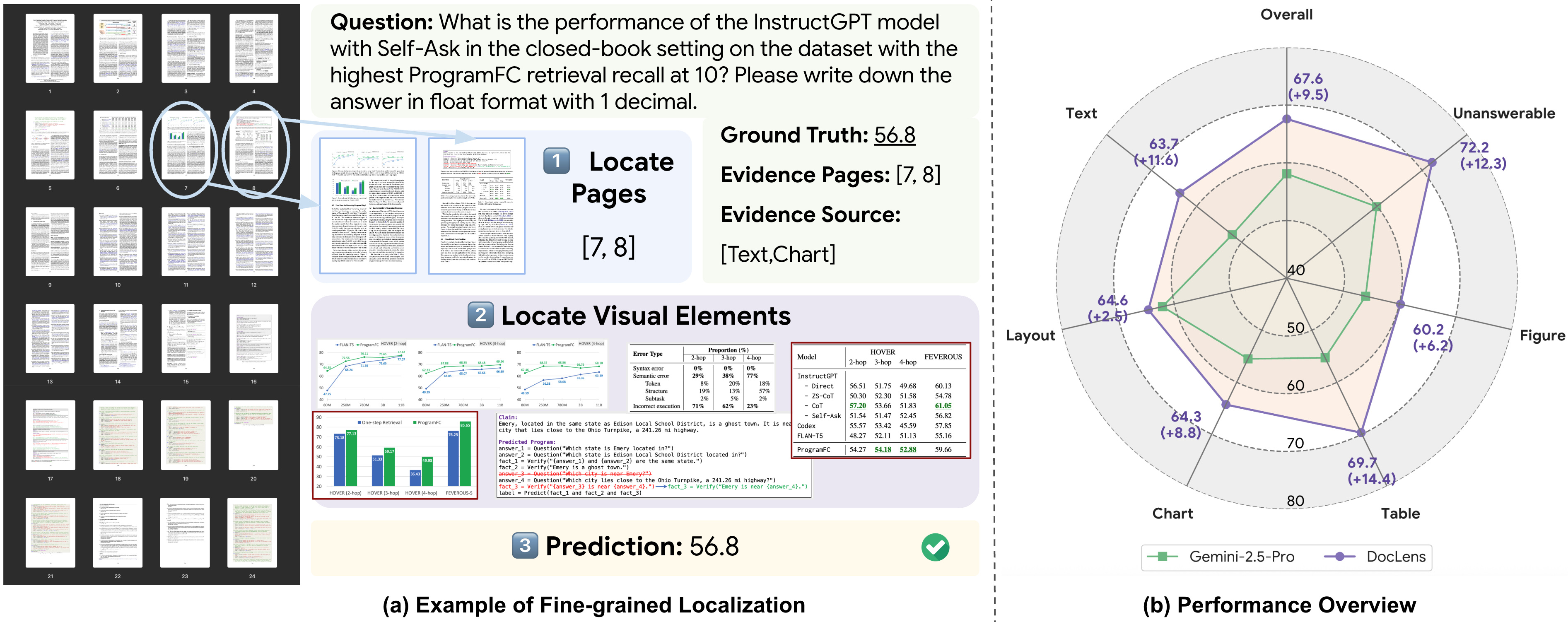}
    \caption{Workflow and performance of our proposed method, \methodname{}. \textbf{(a) }The workflow grounds its answer by navigating from the full document to visual elements (e.g., Text, Chart) within relevant pages. \textbf{(b)} It yields great improvement on MMLongBench-Doc, specifically for understanding visual elements and reducing hallucination.}
    \label{fig:teaser_figure}
\end{figure}

\section{Introduction}

A vast repository of human knowledge is encapsulated in  long visual documents such as financial reports, academic papers, and technical manuals~\citep{liu2025comprehensive}. With information synthesized from various textual and visual elements (tables, charts, figures) distributed throughout the context, these long visual documents are formidably challenging to decipher, even for the most advanced Vision-Language Models (VLMs)~\citep{comanici2025gemini,Claude4,gpt5,Qwen3VL,coreteam2025mimovltechnicalreport,guo2025seed15vltechnicalreport}.

This challenge stems from a fundamental problem: evidence localization. Existing efforts to localizing evidence from long visual documents primarily operate at the page level, either feeding page screenshots to long-context VLMs~\citep{ma2024mmlongbench} or employing vector-based retrieval methods~\citep{han2025mdocagent,cho2024m3docrag}. However, we observe that both approaches perform poorly in recalling evidence pages. On MMLongBench-Doc~\citep{ma2024mmlongbench}, Gemini-2.5-Pro~\citep{comanici2025gemini} only recalls 68\% of evidence pages, while vector-based methods using ColBERT~\citep{santhanam2021colbertv2} and ColPali~\citep{faysse2024colpali} achieve merely 55.3\% Recall@10. This fundamental failure prevents models from producing accurate answers. Moreover, even on the correct page, crucial details within visual elements (e.g., charts, tables) remain obscured in a full-page view, akin to reading a map without a magnifying glass. This dual-level failure in evidence localization—at both the page and element scale—directly fuels model hallucination, causing models to invent responses for over half of unanswerable queries rather than admitting uncertainty on MMLongBench-Doc (Figure~\ref{fig:teaser_figure}b).

In this paper, we propose \methodname{}, a multi-agent framework that overcomes these challenges by strategically leveraging document-parsing tools. Our core component is the \textit{Lens Module}, which zooms into long visual documents like a lens to perform fine-grained evidence localization~(Figure~\ref{fig:teaser_figure}a). It includes a \textit{Page Navigator} agent and an \textit{Element Localizer} agent. The former uses OCR tools to augment VLMs for page-level retrieval, drastically improving recall of evidence pages; the latter employs layout detection and cropping tools to locate visual elements on these retrieved pages for detailed inspection. Following the Lens Module, the Reasoning Module synthesizes the extracted evidence—including page screenshots, text, and cropped visual elements—to formulate a final answer. To ensure both accuracy and reliability, this module employs a ``sampling-adjudication'' process that first proposes a set of potential answers using an \textit{Answer Sampler} agent, and then critically assesses them using an \textit{Adjudicator} agent to select the best candidate.

We evaluate \methodname{} on two challenging benchmarks, MMLongBench-Doc~\citep{ma2024mmlongbench} and FinRAGBench-V~\citep{zhao2025finragbench}. Our method achieves state-of-the-art performance, significantly reduces hallucination, and for the first time surpasses human experts. This breakthrough is driven by the efficacy of our core components: further analysis reveals that our Page Navigator achieves near-perfect evidence page recall (97.3\%), while the Element Localizer dramatically enhances the comprehension of fine-grained visual details. Our main contributions are threefold:

\begin{itemize}[leftmargin=*]
\item A novel, tool-augmented Lens Module that achieves near-perfect page recall and enables fine-grained inspection of visual elements, effectively solving evidence localization.
\item A sampling-adjudication mechanism within the Reasoning Module that effectively mitigates hallucination and improves answer reliability.
\item The establishment of a new state-of-the-art on MMLongBench-Doc and FinRAGBench-V, and for the first time, surpassing human experts.
\end{itemize}
\section{Problem Formulation}

We address the challenge of question answering over long visual documents. A document is a sequence of pages, $\mathcal{D}=\{P_i\}_{i=1}^{N}$, where each page $P_i$ is a screenshot image. From each page, we can extract text $T_i$ and a set of visual elements $\mathcal{V}_i$ (e.g., tables, figures).
Given a question $Q$, the goal is to generate an accurate answer $A$ that is grounded in a specific set of evidence pages $\mathcal{E} \subseteq \mathcal{D}$. This task can be abstractly formulated as learning a function $f$ that maps the document and question to an answer:
\begin{equation}
A = f(\mathcal{D}, Q),
\label{eqn:high_level}
\end{equation}
However, the sheer volume of information in a long document makes a direct mapping challenging to construct. 
We contend that a more principled approach is to decompose the problem into two stages: First, identifying a concise set of relevant evidence from the vast document, and second, generating the answer based on this evidence.

To formalize this, we factorize the function $f$ into two components. First, an extraction function $f_{extract}$ reads through the document to identify a concise evidence set $\mathcal{S}$  relevant to the question:
\begin{equation}
\mathcal{S} = f_{\text{extract}}(\mathcal{D}, Q),
\end{equation}
This evidence set $\mathcal{S}$ contains the necessary pages from $\mathcal{D}$ and the visual and textual elements within these pages. Second, an answer generation function $f_{generate}$ infers the final answer exclusively from this condensed evidence:
\begin{equation}
A = f_{\text{generate}}(\mathcal{S}, Q),
\end{equation}
We therefore model the composite function as:
\begin{equation}
A = f_{\text{generate}}(f_{\text{extract}}(\mathcal{D}, Q), Q),
\end{equation}
The goal is to design and optimize both $f_{extract}$ and $f_{generate}$ to maximize the accuracy of the predicted answer $A$ w.r.t. the ground-truth answer $A^*$.

\section{\methodname Framework}
\begin{figure}[t]
\centering
\includegraphics[width=0.9\textwidth]{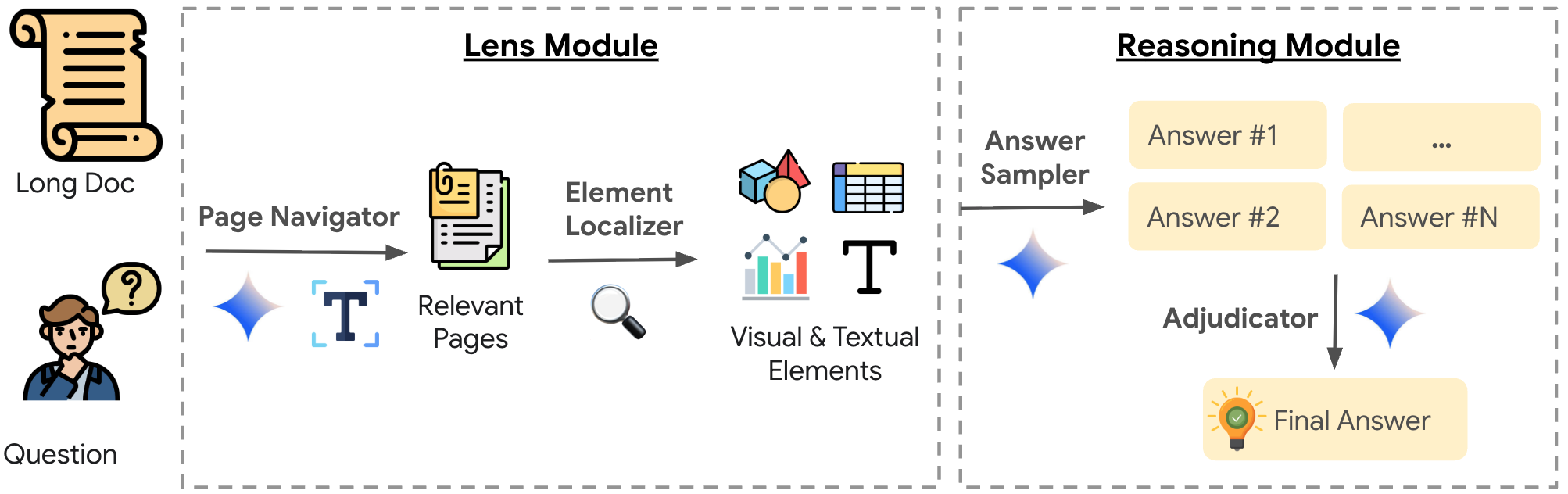}
\caption{Overall workflow of \methodname{}. Given a long visual ducument and a corresponding question, we first apply a \textit{Lens Module} to retrieve relevant pages and locate relevant visual\&textual elements within these pages. We then use a \textit{Reasoning Module} to do in-depth analysis of these elements and provide an accurate answer.}
\label{fig:workflow}
\end{figure}

Figure~\ref{fig:workflow} illustrates the overall workflow of \methodname{}. Our proposed framework consists of two primary components: a Lens Module and a Reasoning Module. Given a long visual document and an associated question, the Lens Module~($f_{\text{extract}}$) is responsible for identifying relevant pages and the key elements within them. Subsequently, the Reasoning Module~($f_{\text{generate}}$) conducts an in-depth analysis of this evidence to generate a precise answer. The prompt templates for all agents and the pseudocode for the entire workflow are presented in Appendices~\ref{sec:app_prompt} and \ref{sec:app_implement}, respectively.

\subsection{Lens Module}
\label{sec:method_lens}
\paragraph{Page Navigator.}
The Lens Module begins with the \textit{Page Navigator} to identify a predicted set of evidence pages, $\mathcal{E}_{\text{pred}}$, from the full document $\mathcal{D} = \{P_i\}_{i=1}^{N}$. First, it uses an OCR tool to extract the text $T_i$ from every page $P_i \in \mathcal{D}$.
\begin{equation}
    T_{i} = \mathsf{OCR}(P_i), \quad i=\{1,2,...,N\},
\end{equation}
To locate potential evidence pages $\mathcal{E}\subseteq \mathcal{D}$, the Page Navigator then prompts an LLM with the question $Q$, all page screenshots and their OCR text (as interleaved input). To ensure comprehensive coverage, this process is repeated $T_e$ times with a temperature $\tau > 0$. Each sampling iteration $j$ generates a candidate page set $\mathcal{E}^{(j)}$:
\begin{equation}
    \mathcal{E}^{(j)} = \mathsf{LLM}_{\text{\textit{PageNav}}}(Q,\{(P_i,T_{i})\}_{i=1}^{N}),
\end{equation}
The final set of predicted pages is the union of all pages identified across these iterations:
\begin{equation}
\mathcal{E}_{\text{\textit{pred}}} = \bigcup\nolimits_{j=1}^{T_e} \mathcal{E}^{(j)},
\end{equation}
In practice, an LLM's finite context window may prevent processing all $N$ pages simultaneously. In such cases, we divide pages into chunks, process them in parallel, and merge the resulting $\mathcal{E}_{\text{pred}}$ sets.

\paragraph{Element Localizer.}
Given the set of predicted pages $\mathcal{E}_{\text{pred}}$ identified by the \textit{Page Navigator}, the \textit{Element Localizer} enriches this set by parsing detailed visual and textual elements.

For each page $P_k \in \mathcal{E}_{\text{pred}}$, its corresponding textual content $T_k$ is available from the prior step.
Concurrently, a layout detection tool identifies the bounding boxes of key visual elements (such as figures, charts, and tables). These elements are then cropped from the page to form a set of focused visual inputs, denoted as $\mathcal{V}_k$:
\begin{equation}
    \mathcal{V}_k = \{ \mathsf{Crop}(P_k, bbox) \mid bbox \in \mathsf{LayoutDetect}(P_k) \},
\end{equation}
where $bbox$ is the bounding box of each visual element.
Then with all predicted evidence pages, we construct the full evidence set $\mathcal{S}$ by collecting tuples of the page screenshot ($P_k$), its extracted text ($T_{k}$), and its cropped visual elements ($\mathcal{V}_{k}$):
\begin{equation}
    \mathcal{S} = \{ (P_k, T_k, \mathcal{V}_k) \mid P_k \in \mathcal{E}_{\text{\textit{pred}}} \}
\end{equation}

\subsection{Reasoning Module}
\paragraph{Answer Sampler.}
The Answer Sampler agent receives the collection of evidence $\mathcal{S}$ extracted by the Lens Module. 
It then integrates all this information to generate a reasoning process $R$ (e.g., a chain-of-thought trace) and a corresponding answer $A$:
\begin{equation}
R, A = \mathsf{LLM}_{\text{\textit{Sampler}}}(Q, \mathcal{S}),
\end{equation}
To generate a diverse set of candidate answers~\citep{wang2023self}, we perform this reasoning process $T_a$ times. The diversity is achieved by strategy with a temperature $\tau>0$. This encourages the model to explore different reasoning paths and wording, yielding $T_a$ distinct reasoning-answer pairs: $\{R_i,A_i\}_{i=1}^{T_a}$.

\paragraph{Adjudicator.}
The final step is managed by the Adjudicator, whose goal is to synthesize the best answer from the $T_a$ candidate answers. It carefully analyzes the reasoning path $R_i$ of each candidate and cross-validates the different approaches to identify the most consistent and logical conclusion, which is then presented as the final answer $A_{final}$:
\begin{equation}
    A_{final}=\mathsf{LLM}_{\text{\textit{Adjud}}}(\{(R_i,A_i)\}_{i=1}^{T_a}).
\end{equation}

\section{Experiments}

In this section, we first introduce our experimental setup~(\S~\ref{sec:exp_setup}), including benchmarks, metrics, tested models and baseline methods. We then demonstrate the overall effectiveness of our \methodname{} on the two selected benchmarks (\S~\ref{sec:exp_main_result_mmlong} \& \S~\ref{sec:exp_main_result_finrag}). For implementation details, please refer to Appendix~\ref{sec:app_implement}.

\subsection{Experimental Setup}
\label{sec:exp_setup}
\paragraph{Benchmarks and Metrics.} We evaluate our method on two challenging benchmarks: MMLongBench-Doc~\citep{ma2024mmlongbench} and FinRAGBench-V~\citep{zhao2025finragbench}.
MMLongBench-Doc tests reasoning over lengthy, multi-domain documents (avg. 49.4 pages) that require integrating scattered information across diverse modalities. Crucially, its dedicated ``Unanswerable'' subset directly evaluates our model's ability to mitigate hallucination. Performance of human experts on this benchmark is reported as 65.8.
FinRAGBench-V is vital for our analysis due to two unique features: its use of documents with dense, newspaper-like layouts~(See Figure~\ref{fig:ana_case_study}), and its support for evaluating visual citation (pinpointing block-level evidence), which provides a direct assessment of our fine-grained localization strategy.
We adhere to the original evaluation protocols: rule-based scoring for MMLongBench-Doc and an LLM-as-a-judge approach for FinRAGBench-V. Further statistics and evaluation details are provided in Appendices~\ref{sec:app_data},\ref{sec:app_mmlong_eval}, and~\ref{sec:app_finrag_eval}.

\begin{table*}[t]
\centering
\footnotesize
\renewcommand{\arraystretch}{1.1}
\setlength\tabcolsep{4pt}
\begin{tabular}{@{}l cccccc|c| ccc|c@{}}
\toprule
\multirow{2}{*}{\textbf{Model}} & \multicolumn{7}{c}{\textbf{MMLongBench-Doc}} & \multicolumn{4}{c}{\textbf{FinRAGBench-V}} \\
\cmidrule(r){2-8} \cmidrule(l){9-12}
& \textbf{TXT} & \textbf{LAY} & \textbf{CHA} & \textbf{TAB} & \textbf{FIG} & \textbf{UNA} & \textbf{ALL} & \textbf{TXT} & \textbf{TAB} & \textbf{CHA} & \textbf{ALL} \\ \midrule
\multicolumn{12}{@{}c}{\textit{Vanilla VLMs}} \\ \midrule
GPT-4o$^{\dag}$       & 46.3 & 46.0 & 45.3 & 50.0 & 44.1 & 20.2 & 42.8 & - & - & - & 37.2 \\
Claude-4-Sonnet       & 50.4 & 49.4 & 50.5 & \underline{57.3} & 43.9 & 59.0  & 53.4 & 36.6 & 20.2 & \underline{51.9} & 33.8 \\
Gemini-2.5-Flash & 44.0 & 53.2 & 46.0 & 43.9 & 48.2 & 56.7 & 49.6 & 49.0 & 41.6 & 41.0 & 43.0 \\
Gemini-2.5-Pro & \underline{52.1} & \underline{62.1} & \underline{55.5} & 55.3 & \underline{54.0} & \underline{59.9} & \underline{58.1} & \underline{62.2} & \underline{55.3} & 50.4 & 54.9 \\ 
o4-mini$^{\dag}$       & - & - & - & - & - & - & - & - & - & - & \underline{62.4} \\ \midrule
\multicolumn{12}{@{}c}{\textit{VLMs Augmented with OCR}} \\ \midrule
Claude-4-Sonnet   &  52.7 & 51.6 & 50.0 & 58.1 &  45.3 & \underline{65.9} & 56.0 & 58.7 & 21.6 & 54.3 & 41.0 \\
Gemini-2.5-Flash & 55.9 & 54.9 & 52.7 & 63.4 & 50.3 & 60.8 & 58.5 & 67.6 & 64.4 & 46.1 & 58.3 \\
Gemini-2.5-Pro & \underline{59.7} & \textbf{\underline{65.3}} & \underline{60.8} & \underline{68.3} & \underline{55.7} & 58.4 & \underline{63.3} & \underline{70.0} & \underline{70.0} & \underline{56.2} & \underline{64.9} \\ \midrule
\multicolumn{12}{@{}c}{\textit{VLM-based Agentic Frameworks}} \\ \midrule
MACT (w/ MiMo-VL-7B)$^{\dag}$  & - & - & - & - & - & - & 47.4 & - & - & - & - \\
M3DocRAG (w/ Qwen2-VL-7B)$^{\dag}$  & 30.0 & 23.5 & 18.9 & 20.1 & 20.8 & 5.8 & 21.0  & - & - & - & - \\
MDocAgent (w/ GPT-4o)$^{\dag}$  & - & - & - & - & - & - & 42.0 & - & - & - & - \\
\multicolumn{12}{@{}l}{\textbf{SimpleDoc}} \\
\quad w/ Claude-4-Sonnet & 52.1 & 53.3 & 58.3 & 62.4 & 46.9 & 66.5 & 58.6 &  59.6 & 68.9 & 54.9 & 61.7 \\
\quad w/ Gemini-2.5-Flash & 45.5 & 57.4 & 49.0 & 51.6 & 45.2 & 66.5 & 53.3 & \textbf{\underline{70.2}} & 56.2 & 53.6 &  58.3 \\
\quad w/ Gemini-2.5-Pro & 48.4 & 54.8 & 55.7 & 56.1 & 52.5 & 59.7 & 56.6 & 67.5 & 64.0 & 60.9 & 63.6 \\
\multicolumn{12}{@{}l}{\cellcolor{gray!15}\textbf{\methodname{} (Ours)}} \\
\quad w/ Claude-4-Sonnet & 59.9 & 58.2 & 54.4 & 63.9 & 55.3 & \textbf{\underline{74.0}} & 63.3 & \textbf{\underline{70.2}} & 66.0 & 60.3 & 64.8 \\
\quad w/ Gemini-2.5-Flash & 59.5 & 61.5 & 54.8 & 66.9 & 59.0 & 73.8 & 64.7 & 69.9 & 71.3 & 64.5 & 68.5 \\
\quad w/ Gemini-2.5-Pro &  \textbf{\underline{63.7}} & \underline{64.6} & \textbf{\underline{64.3}} & \textbf{\underline{69.7}} & \textbf{\underline{60.2}} & 72.2 & \textbf{\underline{67.6}$^{*}$} & 68.9 & \textbf{\underline{74.2}} & \textbf{\underline{67.1}} & \textbf{\underline{70.4}} \\
\bottomrule
\end{tabular}
\caption{Main Results on the MMLongBench-Doc and FinRAGBench-V benchmarks. We report the  accuracy of five types of evidence sources including pure text (TXT), layout (LAY), chart (CHA), table (TAB), and figure (FIG), and on unanswerable (UNA) samples. \textbf{Bold} indicates the best score per column; \underline{underlined} indicates the best per column within each block.\dag{} denotes results reported in the original paper, hence some results are unavailable. $^{*}$ Denotes results surpassing human experts (On MMLongBench-Doc, performance of human experts is 65.8).}
\label{tab:exp_main_results}
\end{table*}

\paragraph{Models and Baselines.} 
We evaluate our proposed agentic framework on three cutting-edge proprietary models: Gemini-2.5-Pro~\citep{comanici2025gemini}, Gemini-2.5-Flash~\citep{comanici2025gemini}, and Claude-4-Sonnet~\citep{Claude4}.
We benchmark its performance against three categories of baselines. The first is the vanilla setting, which uses only page screenshots. The second augments screenshots with OCR text appended to each page, an approach we found particularly effective during our pilot study. The third category comprises existing agentic frameworks: MACT~\citep{yu2025visual}, M3DocRAG~\citep{cho2024m3docrag}, MDocAgent~\citep{han2025mdocagent}, and SimpleDoc~\citep{jain2025simpledoc}. For MACT, M3DocRAG, and MDocAgent, we report the best scores from their original papers. For SimpleDoc, the most recent and best-performing training-free framework, we reproduce results across all proprietary models using our metrics to ensure fair comparison.

\subsection{Main Results on MMLongBench-Doc}
\label{sec:exp_main_result_mmlong}

Table~\ref{tab:exp_main_results} presents our main experimental results. On MMLongBench-Doc, our approach yields substantial performance improvements across all three backbone models. These gains are particularly pronounced for comparatively weaker models, such as Claude-4-Sonnet and Gemini-2.5-Flash, compared to the more powerful Gemini-2.5-Pro.
Notably, our \methodname{} framework enables both Claude-4-Sonnet and Gemini-2.5-Flash to achieve near-human performance. Furthermore, Gemini-2.5-Pro augmented with our method surpasses the human baseline by $\sim2\%$. These results strongly demonstrate the effectiveness of \methodname{}.

Additionally, our method achieves significant improvements on the Unanswerable (UNA) subset, with absolute gains of +8.2\%, +13.0\%, and +13.8\% for Claude-4-Sonnet, Gemini-2.5-Flash, and Gemini-2.5-Pro, respectively. This indicates that our agentic framework effectively mitigates model hallucination, a critical capability for real-world applications.

Finally, we observe that augmenting models with OCR text substantially improves performance across all four backbones compared to the vanilla setting. We attribute this improvement to OCR's effectiveness in facilitating (implicit) evidence page retrieval, as further analyzed in Section~\ref{sec:ana_page_navigator}.

\subsection{Main Results on FinRAGBench-V}
\label{sec:exp_main_result_finrag}

On the FinRAGBench-V benchmark, our framework's superiority is even more pronounced. Compared to the strongest baseline, \methodname{} achieves substantial gains when paired with Claude-4-Sonnet (+3.1\%), Gemini-2.5-Flash (+10.2\%), and Gemini-2.5-Pro (+5.5\%). We hypothesize these larger gains stem from FinRAGBench-V's higher proportion of documents with dense, complex visual layouts (e.g., newspapers).

A closer analysis confirms this hypothesis, revealing that the performance boost is primarily driven by our method's superior handling of visual evidence. On chart-based questions, for instance, \methodname{} elevates the performance of Gemini-2.5-Pro and Gemini-2.5-Flash by absolute margins of +10.9\% and +23.5\% over the strong OCR-augmented baseline. This trend continues for table-based questions, with corresponding gains of +4.2\% and +6.9\%. Collectively, these results demonstrate that as visual complexity increases, the advantage of our fine-grained element localization becomes increasingly critical—a capability we analyze in further detail in Section~\ref{sec:ana_element_localizer}.

\section{Analysis}
\label{sec:analysis}

This section presents a comprehensive analysis of our framework. By default, all experiments are conducted with Gemini-2.5-Pro.
We begin with an ablation study~(\S~\ref{sec:ana_ablation}), which confirms that the Lens Module significantly boosts performance and the Reasoning Module can further reduce hallucination. We then delve deeper into the Lens Module~(\S~\ref{sec:ana_page_navigator} and \S~\ref{sec:ana_element_localizer}), demonstrating through quantitative analysis and case studies how its Page Navigator improves page recall and its Element Localizer enhances visual comprehension by pinpointing specific elements. 
Finally, we demonstrate the framework's efficiency via a hybrid-backbone variant that outperforms baseline with much lower cost (\S~\ref{sec:ana_lowcost}).
We also discussed test-time scaling effect of our method in the Appendix~\ref{sec:app_test_time_scaling}.

\subsection{Ablation on Core Modules}
\label{sec:ana_ablation}

\begin{table}[t]
\centering

\begin{minipage}[b]{0.48\textwidth}  
    \footnotesize  
    \renewcommand{\arraystretch}{1.1}  
    \setlength\tabcolsep{3pt}  
    \centering
    \resizebox{0.99\textwidth}{!}{
    \begin{tabular}{@{}l ccc c@{}}  
        \toprule  
        \multirow{2}{*}{\textbf{Methods}} & \multicolumn{3}{c}{\textbf{MMLong}} & \multirow{2}{*}{\textbf{FinRAG}} \\  
        \cmidrule(r){2-4}  
        & ANS & UNA & ALL &  \\  
        \midrule  
        \methodname{}~(Gemini-2.5-Pro) & 66.4 & \underline{72.2} & \textbf{\underline{67.6}} & \textbf{\underline{70.4}} \\  
        \quad w/o Lens Module & 63.3 & 64.4 & 63.5 & 65.1 \\  
        \quad w/o Reasoning Module & \textbf{\underline{66.6}} & 68.2 & 67.0 & 69.9 \\  
        \midrule  
        \methodname{}~(Gemini-2.5-Flash) & \underline{62.4} & \textbf{\underline{73.8}} & \underline{64.7} & \underline{68.5} \\  
        \quad w/o Lens Module & 58.4 & 68.4 & 60.4 & 60.9 \\  
        \quad w/o Reasoning Module & 62.0 & 71.1 & 63.8 & 67.1\\  
        \bottomrule  
    \end{tabular}  }
    \caption{Ablation study of key modules in our proposed method. \textit{MMLong}, \textit{FinRAG}, \textit{ANS}, \textit{UNS} is short for \textit{MMLongBench-Doc}, \textit{FinRAGBench-V}, \textit{Answerable}, \textit{Unanswerable}, respectively.}  
    \label{tab:exp_ablation_study}  
\end{minipage}
\hfill  
\begin{minipage}[b]{0.48\textwidth}  
    \footnotesize  
    \renewcommand{\arraystretch}{1.1}  
    \setlength\tabcolsep{2pt}  
    \centering
    \resizebox{0.99\textwidth}{!}{
    \begin{tabular}{@{}l|cccc@{}}  
        \toprule  
        \textbf{Setting} & \textbf{\#Pages} & \textbf{Recall} & \textbf{Prec} & \textbf{Final Acc} \\ \midrule  
        Evidence Pages (Oracle) & 1.5 & 100.0 & 100.0 & 69.1 \\ \midrule  
        \multicolumn{5}{@{}l}{\textcolor{gray}{\textbf{Baseline Retrievers}}} \\   
        MDocAgent's Retriever & 13.6 & 71.1 & 7.0 & 49.6 \\  
        SimpleDoc's Retriever & 4.9 & 89.0 & 34.7 & 64.0 \\ \midrule  
        \multicolumn{2}{@{}l}{\methodname{}'s Page Navigator} & & &   \\   
        w/ Gemini-2.5-Pro & 3.5 & \textbf{97.3} & 55.1 & \textbf{67.6}   \\   
        w/ Gemini-2.5-Flash & 3.1 & 95.2 & \textbf{62.0} & 67.1 \\  
        w/ Gemini-2.5-Flash-Lite & 3.2& 90.2 & 60.0 & 64.4 \\   
        \bottomrule  
    \end{tabular}  }
    \caption{Final Accuracy On MMLongBench-Doc with varying retrieval backbones for the Page Navigator. \textit{\#Pages} denotes average number of retrieved pages. \textit{Prec} is short for \textit{Precision}.}  
    \label{tab:ana_retriever}  
\end{minipage}

\end{table}

Table~\ref{tab:exp_ablation_study} ablates the efficacy of our  Lens Module and Reasoning Module.
The ablation settings are as follows: to ablate the Lens Module, we provide the raw screenshot and OCR text directly to the Reasoning Module. To ablate the Reasoning Module, we take the output from the Lens Module (relevant pages and elements) and send it directly to the backbone VLM for answer generation.
The results underscore the critical role of the Lens Module. Its removal leads to a substantial performance drop of 4.1\%  on MMLongBench-Doc and 5.3\% on FinRAGBench-V for Gemini-2.5-Pro, and similar degradation for Gemini-2.5-Flash (7.6\% and 1.4\%, respectively). The Reasoning Module, meanwhile, can further reduce model hallucination. Its absence leads to a noticeable drop in performance on the unanswerable (UNA) queries.

\subsection{Analysis on Page Navigator}
\label{sec:ana_page_navigator}

\paragraph{Page Navigator achieves nearly perfect recall of evidence pages.}

First, we demonstrate the effectiveness of our Page Navigator in terms of retrieving evidence pages. We assess this aspect in two ways: by calculating the recall of retrieved pages against the annotated evidence pages, and by measuring the final accuracy after processing these pages with Element Localizer, Answer Sampler, and Adjudicator. 
As presented in Table~\ref{tab:ana_retriever}, on MMLongBench-Doc, our Page Navigator backboned with Gemini-2.5-Pro achieve a near-perfect recall of 97.3\%, and its final accuracy is only 1.5\% behind using oracle pages. 

\paragraph{Page Navigator outperforms other retrievers.}
We then conduct a comparative analysis by substituting our Page Navigator with retrievers from leading prior work. The first and most prevalent category is vector-based retrievers~\citep{dong2025mmdocir,han2025mdocagent,cho2024m3docrag}, exemplified by MDocAgent~\citep{han2025mdocagent}, which retrieve the top-K most similar pages based on vector representations of the query and individual pages. We use ColBERT~\citep{santhanam2021colbertv2} for textual retrieval and ColPali~\citep{faysse2024colpali} for visual retrieval, combining their top-10 results. More recently, SimpleDoc~\citep{jain2025simpledoc} introduces a refinement through a two-stage pipeline: first retrieving pages via ColQwen2.5, then using an LLM to select evidence pages based on generated summaries. We substitute the original LLM with Gemini-2.5-Pro to maintain consistency with our framework's backbone.

As shown in Table~\ref{tab:ana_retriever}, MDocAgent's vector-based retriever retrieves 13.6 pages on average but achieves low recall on evidence pages, resulting in the poorest final accuracy. SimpleDoc significantly improves both recall and precision, scoring 64.4\% for final accuracy. However, it still underperforms our Page Navigator even when using the weakest backbone. These results validate the effectiveness of our Page Navigator for evidence page retrieval.

\paragraph{OCR significantly enhances page retrieval.}

\begin{wraptable}{r}{0.5\textwidth}
\footnotesize
\renewcommand{\arraystretch}{1.1}
\setlength\tabcolsep{5pt}
\centering
\vspace{-2em}
\begin{tabular}{@{}l ccc c@{}}
\toprule
\multirow{2}{*}{\textbf{Methods}} & \multicolumn{2}{c}{\textbf{MMLong}} & \multicolumn{2}{c}{\textbf{FinRAG}} \\
\cmidrule(r){2-3} \cmidrule(r){4-5}
& Recall & Final Acc & Recall & Final Acc \\
\midrule
\multicolumn{5}{@{}l}{\textcolor{gray}{\textbf{Gemini-2.5-Pro}}} \\
Page Navigator & \textbf{97.3} & \textbf{67.6}  & \textbf{94.3} & \textbf{70.4} \\
\quad w/o Sampling & 95.6 & 66.5 & 89.9 & 69.0 \\
\quad w/o OCR & 87.3 & 58.1 & 53.8 & 50.0 \\ \midrule
\multicolumn{5}{@{}l}{\textcolor{gray}{\textbf{Gemini-2.5-Flash}}} \\
Page Navigator & 95.2 &  64.7 & 90.4 & 68.5 \\
\quad w/o Sampling & 88.0 &  58.1  & 78.3  &  60.2 \\
\quad w/o OCR & 79.3 & 49.6 & 40.3 & 45.6 \\
\bottomrule
\end{tabular}
\caption{Ablation of Page Navigator in terms of retrieving evidence pages (recall) and impact on final performance (accuracy). \textit{MMLong} and \textit{FinRAG} is short for \textit{MMLongBench-Doc} and \textit{FinRAGBench-V}.}
\label{tab:ana_pav_ablation}
\end{wraptable}

We now examine each design choice of our Page Navigator. Table~\ref{tab:ana_pav_ablation} presents ablation results on MMLongBench-Doc and FinRAGBench-V. We observe that both sampling and OCR augmentation improve evidence page recall, which successfully translates into higher final accuracy. Most notably, OCR provides substantial improvements: for Gemini-2.5-Pro, it increases recall by 10.0\% and 40.5\% on MMLongBench-Doc and FinRAGBench-V, respectively; for Gemini-2.5-Flash, the gains are 15.9\% and 50.1\%. This indicates that VLMs are more adept at retrieving relevant information in the textual domain than in the visual domain.

\subsection{Analysis on Element Localizer}
\label{sec:ana_element_localizer}

\paragraph{Element Localizer enhances block-level evidence identification.}

FinRAGBench-V provides 202 cases with human-annotated bounding boxes for relevant blocks on evidence pages. We leverage this subset to examine whether the element localizer improves block-level evidence identification. Specifically, we provide evidence pages to Gemini-2.5-Pro and compare bounding box predictions with and without the element localizer. As shown in Figure~\ref{fig:ana_recall_element_localizer}, the Element Localizer substantially improves block-level performance, increasing precision by 4.9\%, recall by 9.3\%, and F1 score by 6.7\%. (See Appendix~\ref{sec:app_finrag_eval} for detailed calculations ) This also enhances the reliability and traceability of the final output~\citep{ma2024visa}.

\begin{table}[t]
    \centering
    \begin{minipage}[b]{0.48\textwidth}
        \centering
        \includegraphics[width=\linewidth]{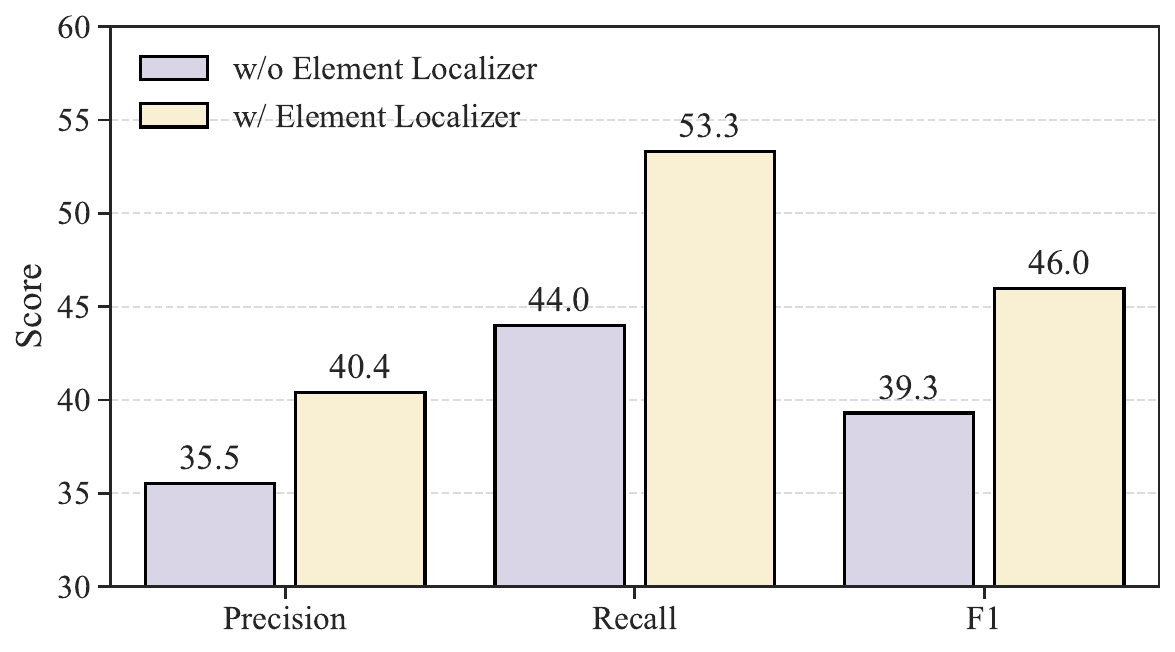}
        \makeatletter\def\@captype{figure}\makeatother\caption{Element Localizer enhances block-level evidence identification.}
        \label{fig:ana_recall_element_localizer}
    \end{minipage}
    \hfill
    \begin{minipage}[b]{0.48\textwidth}
        \centering
        \includegraphics[width=\linewidth]{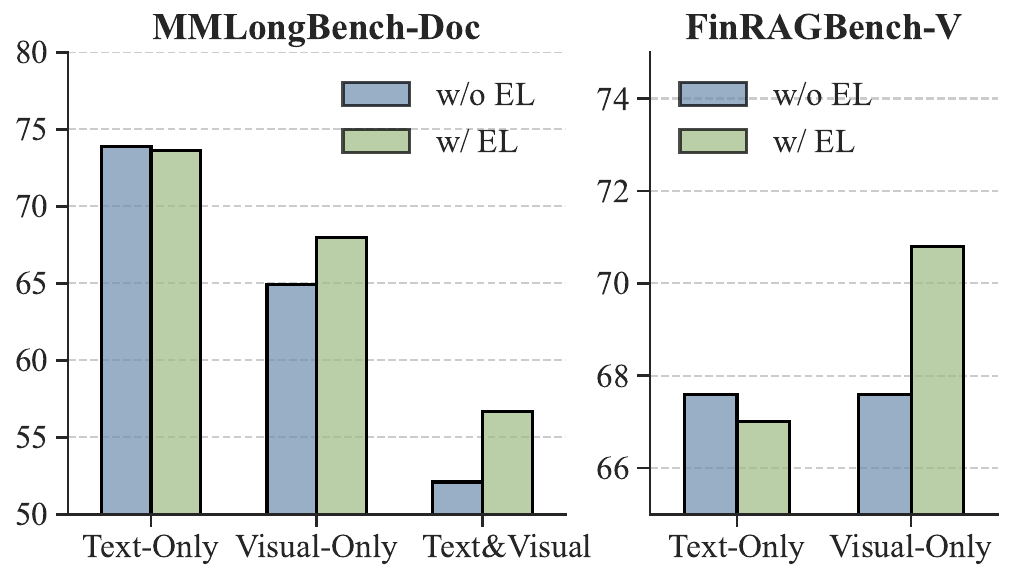}
        \makeatletter\def\@captype{figure}\makeatother\caption{Effectiveness of Element Localizer~(\textit{EL}) on different evidence sources.}
        \label{fig:ana_element_localizer}
    \end{minipage}
\end{table}

\paragraph{Effectiveness on visual-centric queries.}
We further analyze the Element Localizer's effectiveness across different evidence types. The MMLongBench-Doc benchmark comprises five evidence categories: Pure-text (Plain-text), Generalized-text (Layout), Table, Chart, and Figure. We partition samples into three distinct sets based on their evidence sources: \textit{Text-Only}, containing evidence exclusively from the first two categories; \textit{Visual-Only}, containing evidence solely from the latter three categories; and \textit{Text\&Visual}, containing evidence from both domains.
We compare Final Scores with and without the Element Localizer across these three data splits, with results presented in Figure~\ref{fig:ana_element_localizer}. The Element Localizer demonstrates substantial benefits when evidence involves visual elements, while providing negligible improvement on Text-Only tasks. This pattern holds consistently on the FinRAGBench-V benchmark, reinforcing this finding.

\begin{figure}[t]
    \centering
    \includegraphics[width=0.9\textwidth]{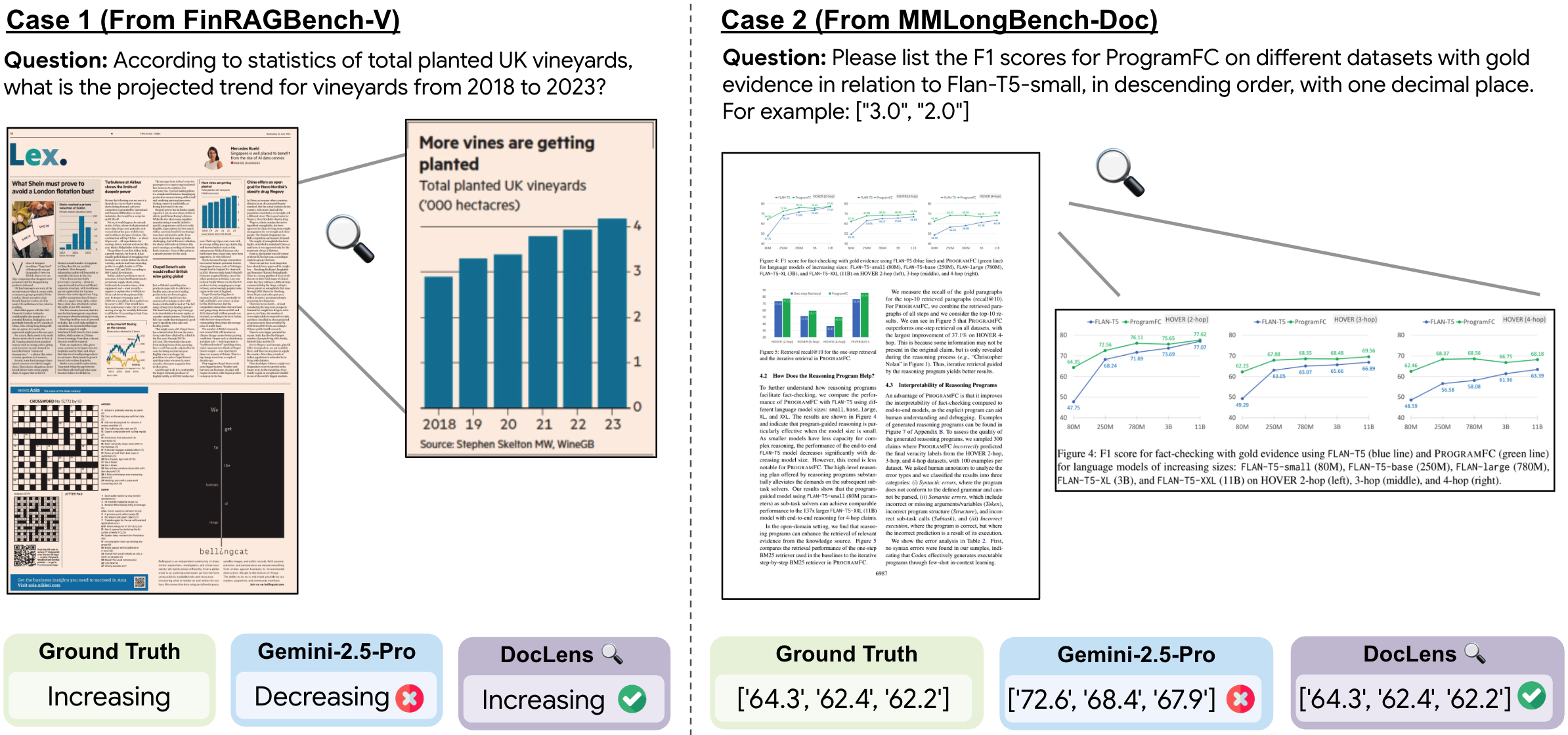}
    \caption{Cases demonstrating the effectiveness of Element Localizer.}
    \label{fig:ana_case_study}
\end{figure}

\paragraph{Case Study.} Figure~\ref{fig:ana_case_study} presents two cases that highlight the effectiveness of the Element Localizer. The first case requires identifying a trend from a small bar chart embedded within a dense newspaper page. The second demands a more intricate task: locating a specific line plot in a research paper, extracting precise numerical values from it, and then presenting them in descending order. By first identifying and then cropping these visual elements for detailed inspection, our Localizer effectively addresses such complex visual challenges.

\subsection{Hybrid Backbones for Cost Efficiency}
\label{sec:ana_lowcost}

Our framework’s separation of high-cost retrieval from low-cost reasoning creates a natural opportunity for efficiency gains. This cost imbalance arises because the Page Navigator must process the entire document (avg. 49.4 pages on MMLongBench-Doc), while the Reasoner only analyzes the retrieved pages (avg. 3.5).
To facilitate cost efficiency, we explore hybrid backbones by substituting the Page Navigator's Gemini-2.5-Pro backbone with cheaper alternatives: Gemini-2.5-Flash and Gemini-2.5-Flash-Lite\footnote{We refer to \url{https://ai.google.dev/gemini-api/docs/pricing}. Gemini-2.5-Pro / Flash / Flash-Lite costs \$1.25 / \$0.3 / \$0.1 per million input tokens, respectively.}. As shown in Table~\ref{tab:ana_retriever}, using Gemini-2.5-Flash (67.1\%) and even the lightweight Gemini-2.5-Flash-Lite (64.4\%) for retrieval both outperform the vanilla Gemini-2.5-Pro baseline (63.3\%). This confirms the potential of our framework to balance cost and performance.

\section{Related Work} \label{sec:related_work}
\paragraph{Visual Document Understanding.}
Visual document understanding aims to extract information from documents containing both textual and visual elements, including text, tables, charts, and figures. Early efforts primarily focus on understanding short, single-page visual documents, establishing foundational benchmarks such as DocVQA~\citep{mathew2021docvqa}, ChartQA~\citep{masry2022chartqa}, and SlideVQA~\citep{tanaka2023slidevqa}. With recent advances in VLMs, models have achieved strong performance on these benchmarks, prompting the research community to shift toward two more challenging directions. The first involves multi-document scenarios~\citep{dong2025mmdocir,zhao2025finragbench,cho2024m3docrag}, which focus on retrieving relevant documents from a corpus and performing retrieval-augmented generation, as exemplified by ViDoRAG~\citep{wang2025vidorag}, M3DocRAG~\citep{cho2024m3docrag}, and VRAG-RL~\citep{wang2025vrag}. The second setting involves single, long-document comprehension~\citep{ma2024mmlongbench,deng2024longdocurl,zou2024docbenchbenchmarkevaluatingllmbased,dong2025benchmarking}, challenging models to process cohesive but extensive visual documents~\citep{yu2025visual,han2025mdocagent,jain2025simpledoc}. Our work situates within this latter context.

\paragraph{Evidence Localization in Visual Document Understanding.}
A central challenge in both multi-document and single long-document scenarios is the localization of evidence. Conventional approaches~\citep{tanaka2025vdocrag,yu2024visrag} primarily rely on vector-based models~\citep{santhanam2021colbertv2,faysse2024colpali,wang2022text,bge-m3,gunther2023jina,zhu2024longembed} to retrieve top-K pages by combining textual and visual features. However, embedding models often struggle to capture complex reasoning relationships~\citep{hongjinbright}, resulting in suboptimal page recall. A recent improvement, SimpleDoc~\citep{jain2025simpledoc}, employs an iterative pipeline that retrieves pages via vector-based models, uses an LLM to select evidence pages from generated summaries, and refines the query based on missing information, repeating until sufficient information is gathered. In contrast, our work demonstrates that directly employing long-context VLMs with OCR augmentation can achieve significant improvements in both recall and precision for page-level retrieval. Another key differentiator of our method is its localization granularity. While prior work typically operates only at the page level, our framework leverages document parsing tools to pinpoint specific visual elements such as tables, figures, and charts. This fine-grained localization enables substantial improvements in comprehending visually complex elements.

\paragraph{Agentic Frameworks for Long Context Modeling} Fueled by recent LLM advancements, agent-based systems are increasingly prominent for their ability to handle complex reasoning tasks via multi-role collaboration and tool usage~\citep{yao2022react,song2023restgpt,tran2025multi,wu2024autogen}. In the realm of long-context modeling, agentic frameworks~\citep{zhang2025agentic,li2025search,ouyang2025reasoningbank,chen2023walking,zhou2023recurrentgpt,zhao2024longrag,edge2024local,yang2025deep} offer a flexible alternative to extending context windows and processing target tasks via end-to-end training~\citep{zhu2023pose,chen2023extending,peng2023yarn}. These frameworks generally fall into two categories: retrieval-augmented generation and memory-augmented generation. The first category~\citep{li2025search,han2025mdocagent,zhao2024longrag,edge2024local,yang2025deep} focuses on first retrieving relevant information followed by generating an answer based on these pieces. The second category~\citep{zhang2025agentic,ouyang2025reasoningbank,chen2023walking,zhou2023recurrentgpt}, in contrast, first compresses long context segments into summaries or more abstract `memory'—often using divide-and-conquer or on-the-fly approaches—and then directs the model to answer questions based on this compressed memory. Our method extends the first category of agentic frameworks by leveraging existing document parsing tools to perform more fine-grained analysis and localization of relevant elements within the document.

\section{Conclusion}

In this paper, we introduced \methodname{}, a tool-augmented multi-agent framework that addresses critical challenges in long visual document understanding: evidence localization. Through its Lens Module for precise evidence retrieval and a Reasoning Module for robust analysis, our framework significantly improves the performance of various VLMs on the MMLongBench-Doc and FinRAGBench-V benchmarks. Notably, \methodname{} with Gemini-2.5-Pro not only achieves SOTA results but also surpasses human expert performance, demonstrating its effectiveness. 
\section*{Limitations}
While this paper makes substantial progress in evidence localization for long visual document understanding and achieves state-of-the-art performance on two challenging benchmarks, several limitations remain.

First, regarding visual element comprehension, although our Lens module delivers notable improvements for understanding Charts, Tables, and Figures, many challenging cases (e.g., See Appendix~\ref{sec:app_hard_cases}) persist that cannot be adequately addressed through simple ``zooming-in'' strategies. Effectively handling these cases requires either designing dedicated agentic frameworks tailored to specific visual element types or advancing the fundamental perception capabilities of backbone LLMs.

Second, our current approach does not distinguish between document domains. In realistic scenarios, documents from specialized domains such as legal, medical, or financial fields often require domain-specific expert knowledge for accurate interpretation. Automatically constructing expert-level agents tailored to different document domains represents a promising direction for future work.
\section*{Acknowledgement}

We thank all members of Google Cloud AI Research for their valuable
support during the project.


\bibliographystyle{abbrvnat}
\nobibliography*
\bibliography{ref_bak,custom}

\clearpage

\appendix
\onecolumn
\clearpage

\section{Prompt Templates for our Agentic Framework.}
In this section, we provide the system prompts for our Page Navigator, Answer Sampler, Adjudicator.
\label{sec:app_prompt}

\begin{promptbox}[breakable]{Prompt used for the Page Navigator}
## ROLE
You are an expert AI assistant specializing in multimodal long document understanding. Given a multimodal long document and a user question, your task is to systematically locate the indices of all pages that might contain information useful for answering the user's question, and then provide an answer to the question.

## Follow these instructions carefully:
- Core Objective: Your primary goal is to identify all pages relevant to the question. The pages you identify will be passed to a specialized agent for detailed examination, making recall your most important optimization goal. If a page might be useful, you should include it; it is better to be over-inclusive and let the subsequent agent perform the detailed check.
- Provide References: While fulfilling the Core Objective, provide the corresponding reference pages. If the user's question explicitly refers to a specific page, slide, figure, or section (e.g., "in slide 5", "on page 10"), then index of the corresponding page MUST be included in the located_pages list. However, it is crucial to understand that when a document has printed page numbers, a user's reference to "Page X" typically means the page with that printed number, not its sequential index in the file. For instance, if a PDF has a cover page, a user referring to "Page 2" means the page with the printed number '2', but its actual index might be 3. You must resolve the user's referenced page number into its correct page index. Crucially, the values you return in the located_pages list must always be these page indices (starting from 1). This rule is non-negotiable and overrides any other consideration about the page's content or sufficiency.
- Rules of numerical answers:
    - If the user asks for an absolute number (e.g., with questions like "How many...?"), you must first attempt to locate the number directly. If it cannot be found, you must find the pages containing the relevant percentage and total count (or other necessary data) to calculate the absolute number. If the calculated absolute number for discrete entities (e.g., people, companies, objects) is a decimal, you must round it to the nearest whole number.
    - If the user asks for a percentage (or proportion), you must first attempt to locate the percentage directly. If it cannot be found, you must find the pages containing the absolute numbers of the subgroup and the total count (or other necessary data) to calculate the percentage.
    - If the user's question is ambiguous and does not explicitly specify a number or percentage (e.g., "What's the gap between...?"), you must default to providing the absolute value. If you can only find relative values (percentages) in the chart, you must make every effort to find a total number within the provided context to calculate the absolute value. Only return the relative value as a last resort if a total number cannot be found, and explain that you cannot find total number in this case.

## Output Format:
Your entire response MUST be a single, valid JSON object and nothing else. Do not wrap it in markdown code blocks or add any other text. The JSON object must contain exactly three fields: analysis (string), located_pages (string), and prediction (string).
- analysis field: Briefly explain your thought process. Describe how you located the answer within the document, which pages, tables, or figures you referenced, and how you connected the information to the question.
- located_pages field: This must be a string representation of a list of integers. Page indices start at 1. If relevant pages are found, it should look like this: "[3, 10, 12]". If no pages contain relevant evidence, it MUST be an empty list: "[]". Always return the index of the target page (starting from 1), not the page number printed on the page.
- prediction field: This must be a string containing the direct answer to the user's question.
\end{promptbox}

\newpage
\begin{promptbox}[breakable]{Prompt used for the Answer Sampler}
## ROLE
You are an expert AI assistant specializing in multimodal long document understanding. Your task is to carefully analyze the provided page images (which may contain text, figures, tables, and other content) and provide a precise answer to the user's question. Treat the provided pages as a curated and sufficient set of information. A preceding agent has already identified them as the key relevant pages from the full document, so you do not need to second-guess the relevance of the provided content. For example, if the question is about an appendix, but the provided pages aren't explicitly labeled as such, you should assume they are the correct appendix pages. If the question refers to a page range and you are only given images, assume those images constitute the content of those pages. If the question asks for a specific item (e.g., the "5th FAQ") and you are shown only one, treat that as the target item. Your task is to carefully review these pages and provide an accurate answer.

## Follow these instructions carefully:
- Core Objective: Your primary goal is to accurately and concisely answer the user's question based on the content of the provided document pages.
- Rules of numerical answers:
    - If the user asks for an absolute number (e.g., with questions like "How many...?"), you must first attempt to locate the number directly. If it cannot be found, find the relevant percentage and total count (or other necessary data) to calculate the absolute number. If the calculated absolute number for discrete entities (e.g., people, companies, objects) is a decimal, you must round it to the nearest whole number.
    - If the user asks for a percentage (or proportion), you must first attempt to locate the percentage directly. If it cannot be found, find the absolute numbers of the subgroup and the total count (or other necessary data) to calculate the percentage.
    - If the user's question is ambiguous and does not explicitly specify a number or percentage (e.g., "What's the gap between...?"), you must default to providing the absolute value. If you can only find relative values (percentages) in the chart, you must make every effort to find a total number within the provided context to calculate the absolute value. Only return the relative value as a last resort if a total number cannot be found, and explain that you cannot find total number in this case.
- Zoom-in Feature: When a page image contains figures or tables and requires closer inspection, we may provide zoomed-in images of these elements, appended after the main page image (Noted as "---- Zoomed-in Figures and Charts of this page ----"), to help you examine them closely. We will also extract text from the page image into Markdown format. Note: For questions related to page layout, you must refer to the original page image itself, not the zoomed-in images or the Markdown text, as they may lose layout information. For instance, if asked for the first figure on the page, you should consult the full page image to determine its order, not the sequence of the provided zoomed-in images.
- Page Numbering: Page numbers in the user's question typically refer to the number printed on the page image, not the page's index in the document file. For example, if a PDF's first page is the cover and the third page is the first page of content (labeled "Page 1"), a user's question about "page 1" refers to that third page. Similarly, when asked to provide a page number, you should return the printed page number from the image. Only return the page index if no number is printed on the page.
- Rule of faithfulness: Be faithful. If the provided pages do not contain sufficient information to answer the user's question, you should answer `Not answerable`. For example, if the user asks for a man in green shirts, but there are only man in red shirts in the provided pages, you should answer `Not answerable`; if the user asks for the boy playing badminton, but there are only boys playing football in the provided pages, you should answer `Not answerable`; if the user asks for a certain year's data but the provided pages only contain data for other years, you should answer `Not answerable`; if the user asks for the color of a certain object but the provided pages do not contain that object, you should answer `Not answerable`. 

## Output Format:
Your entire response MUST be a single, valid JSON object and nothing else. Do not wrap it in markdown code blocks or add any other text. The JSON object must contain exactly two fields: analysis (string), and prediction (string).
- analysis field: Briefly explain your thought process. Describe how you located the answer within the document, which pages, tables, or figures you referenced, and how you connected the information to the question.
- prediction field: This must be a string containing the direct answer to the user's question.
\end{promptbox}

\newpage
\begin{promptbox}[breakable]{Prompt used for the Answer Adjudicator}
## Role:
You are an expert AI assistant specializing in multimodal long document understanding. Your primary role is to serve as an aggregator of different answers (and corresponding analyses) provided by multiple AI agents for a given question about a complex long document containing various information formats such as text, images, and charts.

## Follow these instructions carefully:
- Core Objective: Your ultimate goal is to accurately and concisely answer the user's question based on the content of the provided document pages. You will be presented will several answers and analyses from different agents, and you must determine which answer is the most appropriate by evaluating the reasoning behind each one.
- Serving as a judge, not a executor. Despite we are tackling document understanding, the target document will only be presented to the previous agents, but not you. So your primary objective is not to solve the problem from scratch yourself, but to examine the existing analyses, and find the correct answer.
- Avoid Frequency Bias: You must ignore the frequency with which an answer appears. An answer being repeated by multiple agents does not make it correct. Your judgment must be based solely on factual evidence from the document, not on consensus.
- Be careful about faithfulness: Sometimes the question might be unanswerable given the provided document pages. In this case, "Not answerable" should be the desired answer. However, not all agents will be aware of this. Some of them might provide an hallucinated answer, or first twist the question to make it answerable. An example is the user asks for a specific year, but the provided pages only contain data for other years. In this case, some agents might answer with the closest year. Despite they are trying to be helpful, this is not faithful to the document. Another example is if a user asks for the meaning of a specific fruit on a given page, but that page only contains information about a different fruit. Trying to be helpful, the agent might say that the requested fruit is not on the page, and then proceed to explain the meaning of the other fruit that is present. In such cases, the desired answer must still be "Not answerable". It is your duty to indentify such cases, and choose "Not answerable" as the final answer.
- Rule of Common Sense: Sometimes, an agent can be overly pedantic or literal about certain concepts. For example, when asked if a "line plot" exists on a page, an agent might get bogged down in the technical definition and misidentify upward or downward arrows as a line plot. This clearly defies common sense. In reality, the user is an ordinary person. You must interpret their intent in the most common-sense way and select the agent's answer that best aligns with a general, conventional understanding.

## Input Format
You will first be provided with the question, and then a list of Agent responses in the following format:

**Question:**
[The exact question that was asked will be stated here]

**List of Agent Analyses and Answers:**
Agent 1
Analysis: [The reasoning process provided by Agent 1]
Answer: [The final answer provided by Agent 1]
Agent 2
Analysis: [The reasoning process provided by Agent 2]
Answer: [The final answer provided by Agent 2]
Agent 3
[...]

## Output Format:
Your entire response MUST be a single, valid JSON object and nothing else. Do not wrap it in markdown code blocks or add any other text. The JSON object must contain exactly two fields: analysis (string), and prediction (string).
- analysis field: Insert your detailed meta-analysis here. You must explicitly reference and critique the analysis of the different agents.
- prediction field: Insert the exact text of the correct agent answer here, with no prefix
\end{promptbox}

\newpage
\section{Dataset Statistics}
\label{sec:app_data}

Statistics of MMLongBench-Doc and FinRAGBench-V is presented in Table~\ref{tab:data_statistics}. MMLongBench-Doc comprises documents from 7 various domains, including Research report/Introduction, Tutorial/Workshop, Academic Paper, Guidebook, Brochure, Administration/Industry file, Financial Report. FinRAGBench-V, on the other hand, focuses sole on financial reports.

\begin{table}[h!]
\centering
\footnotesize
\renewcommand{\arraystretch}{1.1}
\setlength\tabcolsep{14pt}
\begin{tabular}{l|cc}
\toprule
\multirow{2.5}{*}{\textbf{Statistics}}  &  \multicolumn{2}{c}{\textbf{Dataset Name}} \\  \cmidrule(l){2-3}
& MMLongBench-Doc & FinRAGBench-V \\ \midrule
\textbf{Documents} & 135 & 301 \\
- Average/Medium pages & 47.5 / 28 & 76.1 / 57.0 \\
- Average/Medium words & 8,393 / 5,743 & 36,026 / 16,329 \\ \midrule
\textbf{Total question} & 1,082 & 1,394 \\
- Single-page question & 494 (45.7\%) & 1,218 (87.4\%) \\
- Cross-page question & 365 (33.7\%) & 178 (12.6\%) \\
- Unanswerable question & 223 (20.6\%) &  - \\ \midrule
\textbf{Evidence source} & & \\
- Pure-text & 305 (35.5\%) & 302 (21.7\%) \\
- Layout & 119 (13.9\%) & - \\
- Table & 218 (25.4\%) & 573 (41.1\%) \\
- Chart & 178 (20.7\%) & 519 (37.2\%) \\
- Image & 304 (35.4\%) & - \\ \midrule
Avg. / Max. question words & 16.2 / 54 & 35.8 / 108\\
Avg. / Max. answer words & 2.1 / 66 & 23.4 / 174 \\
\bottomrule
\end{tabular}
\caption{Statistics of benchmarks}
\label{tab:data_statistics}
\end{table}

\section{Implementation Details}

\label{sec:app_implement}

\subsection{Implementing \methodname{}}

\paragraph{Document parsing tools.}
We employ MinerU~\citep{niu2025mineru25decoupledvisionlanguagemodel}, a recently proposed document parsing tool, to perform OCR, layout detection, and cropping.
\paragraph{Sampling.}
Most mainstream API interfaces, including vLLM~\footnote{\url{https://docs.vllm.ai/en/v0.6.4/dev/sampling_params.html}}, OpenAI~\footnote{\url{https://platform.openai.com/docs/api-reference/chat/create\#chat-create-n}}, and Google GenAI SDK~\footnote{\url{https://googleapis.github.io/python-genai/genai.html\#genai.types.GenerationConfig.candidate_count}}, support generating n independent candidate responses for a single input. The cost structure is: input token cost $\times$ 1 + total output token cost across all candidates. Compared to invoking the API nn
n times sequentially, this approach is substantially more efficient in both time and cost. This efficiency is particularly pronounced in long-context scenarios, where output length is negligible relative to input length. Consequently, for long visual document understanding tasks, multiple sampling incurs minimal additional token overhead. The sole exception is Anthropic's API, which does not support this functionality; therefore, for experiments involving Claude, we invoke the API N times sequentially. In our implementation, both the retriever sampling count $T_e$
and the answer sampler sampling count $T_a$ are set to 8, with a temperature $\tau=0.7$.

\paragraph{Page Navigator.} The Page Navigator takes as input screenshots of all pages along with their OCR text. In practice, an LLM's finite context window may prevent processing all $N_p$ pages simultaneously for extremely long documents. To address this, we employ a chunking strategy that divides pages into chunks of $K$ pages each, processes them in parallel with the Page Navigator, and merges the identified pages. In our experiments, this strategy is not required for Gemini-2.5-Pro and Gemini-2.5-Flash, both of which support a 1M token context window. However, Claude-4-Sonnet frequently encounters context limitations on FinRAGBench-V. For this model, we set $K=50$, processing 50 pages per iteration. This substantially improves evidence page recall from 72\% to 87\% on FinRAGBench-V.

\paragraph{Miscellaneous.} Some APIs impose restrictions on image resolution. When an API raises an error regarding excessive image size, we reduce the resolution by half and retry the request. In the VLMs Augmented with OCR setting, if context limit is exceed, we discard the text and roll back to the image only setting.

\paragraph{Pseudo Code.} Algorithm~\ref{alg:workflow} presents overall workflow of our \methodname{} framework.

\subsection{Reproducing SimpleDoc}
\label{sec:app_implement_rep_simpledoc}
SimpleDoc~\citep{jain2025simpledoc} is the most recent training-free agentic framework for long visual document understanding. By optimizing the page retriever, it significantly improves upon previous methods including MDocAgent~\citep{han2025mdocagent} and M3DocRAG~\citep{cho2024m3docrag}. Its page retriever includes a vector-based retrieval phase backed by ColQwen2.5, and a summary-augmented reranking phase backed by Qwen3-30B-A3B~\citep{yang2025qwen3technicalreport}. The selected pages, along with the question, are then fed into Qwen2.5-VL-32B-Instruct for answer generation. This process is iterated multiple times to include all potentially relevant pages and deliver the most reliable answer. Finally, it uses GPT-4o to evaluate the consistency between the generated answer and ground truth.

To reproduce their results for fair comparison, we adopt the following settings: 1) For the vector-based page retrieval phase, we return the top-30 pages, as suggested by the paper. 2) We replaced all generative backbones (Qwen3-30B-A3B, Qwen2.5-VL-32B-Instruct) with Gemini-2.5-Flash, Gemini-2.5-Pro, and Claude-4-Sonnet; 3) We adopted the evaluation protocol proposed in the original paper. Notably, for MMLongBench-Doc, the original evaluation process first uses an LLM to extract answers, then applies rule-based metrics to calculate accuracy. Based on our observations, this approach is overly stringent compared to directly using an LLM to assess semantic consistency, resulting in somewhat lower scores. For reference, when using LLM-based evaluation, our \methodname{} with Gemini-2.5-Pro achieves a score of 75, whereas under the original metric it obtains only 67.6. Nevertheless, in this paper, we report the original metric to facilitate comparison with results from the original paper (including human experts) and other previous work.

\begin{algorithm}[H]
\caption{Overall Workflow of \texttt{\methodname}}
\label{alg:workflow}
\begin{algorithmic}[1]
\Require A visual document $\mathcal{D}=\{P_i\}_{i=1}^{N}$; A question $Q$ in natural language.
\Ensure The final answer $A_{final}$.

\State \Comment{\textbf{Lens Module}: Extract relevant evidence from the document}
\State $\mathcal{E}_{pred} \leftarrow \emptyset$ \Comment{$\triangleright$ Initialize the set of predicted evidence pages.}
\For{$i=1$ to $N$}
    \State $T_i \leftarrow \mathsf{OCR}(P_i)$ \Comment{$\triangleright$ Extract text from each page.}
\EndFor

\State \Comment{\textbf{Page Navigator}: Identify relevant pages}
\For{$j=1$ to $T_e$} \Comment{$\triangleright$ Sample multiple times for comprehensive coverage.}
    \State $\mathcal{E}^{(j)} \leftarrow \mathsf{LLM}_{\text{\textit{PageNav}}}(Q,\{(P_i,T_{i})\}_{i=1}^{N})$
    \State $\mathcal{E}_{pred} \leftarrow \mathcal{E}_{pred} \cup \mathcal{E}^{(j)}$
\EndFor

\State \Comment{\textbf{Element Localizer}: Extract visual and textual elements from predicted pages}
\State $\mathcal{S} \leftarrow \emptyset$ \Comment{$\triangleright$ Initialize the full evidence set.}
\ForAll{page $P_k \in \mathcal{E}_{pred}$}
    \State $T_k \leftarrow \text{text extracted for } P_k \text{ in line 4}$
    \State $bboxes \leftarrow \mathsf{LayoutDetect}(P_k)$ \Comment{$\triangleright$ Identify bounding boxes of visual elements.}
    \State $\mathcal{V}_k \leftarrow \emptyset$
    \ForAll{$bbox \in bboxes$}
        \State $\mathcal{V}_k \leftarrow \mathcal{V}_k \cup \{ \mathsf{Crop}(P_k, bbox) \}$ \Comment{$\triangleright$ Crop visual elements from the page.}
    \EndFor
    \State $\mathcal{S} \leftarrow \mathcal{S} \cup \{ (P_k, T_k, \mathcal{V}_k) \}$ \Comment{$\triangleright$ Aggregate page screenshot, text, and visuals.}
\EndFor

\Statex
\State \Comment{\textbf{Reasoning Module}: Generate the final answer from the evidence}
\State \Comment{\textbf{Answer Sampler}: Generate multiple candidate answers}
\State $\{R_i,A_i\}_{i=1}^{T_a} \leftarrow \emptyset$ \Comment{$\triangleright$ Initialize a set for reasoning-answer pairs.}
\For{$i=1$ to $T_a$} \Comment{$\triangleright$ Generate diverse reasoning paths.}
    \State $R_i, A_i \leftarrow \mathsf{LLM}_{\text{\textit{Sampler}}}(Q, \mathcal{S})$
\EndFor

\State \Comment{\textbf{Adjudicator}: Synthesize the final answer}
\State $A_{final} \leftarrow \mathsf{LLM}_{\text{\textit{Adjud}}}(\{(R_i,A_i)\}_{i=1}^{T_a})$ \Comment{$\triangleright$ Select the most consistent and logical conclusion.}
\State \Return $A_{final}$
\end{algorithmic}
\end{algorithm}

\subsection{Evaluation on MMLongBench-Doc}
\label{sec:app_mmlong_eval}
1. For MMLongBench-Doc, we manually fixed some annotations about evidence pages.

\begin{promptbox}[breakable]{Prompt used for Answer Extraction on MMLongBench-Doc}
Given the question and analysis, you are tasked to extract answers with required formats from the free-form analysis. 
- Your extracted answers should be one of the following formats: (1) Integer, (2) Float, (3) String and (4) List. If you find the analysis and the question can not be answered from the given documents, type "Not answerable". Exception: If the analysis only tells you that it can not read/understand the images or documents, type "Fail to answer".
- Please make your response as concise as possible. Also note that your response should be formatted as below:
```
Extracted answer: [answer]
Answer format: [answer format]
```

Please read the following example, then extract the answer from the model response and type it at the end of the prompt. 

---
Question: List the primary questions asked about the services in this report.
Analysis:  The primary questions asked about the services in the report for The Limes Residential Home are:\n\n1. Is the service safe?\n2. Is the service effective?\n3. Is the service caring?\n4. Is the service responsive?\n5. Is the service well-led?
Extracted answer: ['Is the servife safe?', 'Is the service effective', 'Is the serve caring?', 'Is the service responsive?', 'Is the service well-led?']
Answer format: List

---
Question: How many regulations of the HSCA 2008 are breached in all according to this report?
Analysis: According to the report, the provider breached 10 Health and Social Care Act 2008 (Regulated Activities) Regulations in total. Here are the specifics:\n\n1. Regulation 13: Safeguarding service users from abuse and improper treatment\n2. Regulation 12: Safe care and treatment\n3. Regulation 18: Staffing\n4. Regulation 11: Need for consent\n5. Regulation 10: Dignity and respect\n6. Regulation 9: Person-centred care\n7. Regulation 17: Good governance\n8. Regulation 18 (CQC Registration Regulations 2009): Notification of other incidents\n9. Regulation 18: Failure to maintain an accurate and up-to-date care plan\n10. Regulation 11: Failure to implement the Mental Capacity Act 2005 code of practice effectively\n\nThese breaches involve issues concerning staffing, safeguarding, medicines management, dignity and respect, consent, care planning, governance, and failure to notify the CQC of incidents.
Extracted answer: 10
Answer format: Integer

---
Question: According to the survey that is the percentage of Chinese who are paying more or about the same attention to politics after Trump's election?
Analysis: The survey provided does not specify the percentage of Chinese individuals specifically who are paying more or about the same attention to politics after Trump's election. The report focuses primarily on American demographics and does not include specific details about the Chinese population in relation to this question. If you need information about a different demographic or a summary of the findings from the American demographic, I can certainly help with that!
Extracted answer: Not answerable
Answer format: String

---
Question: How many quotations from male respondent over 50 years old are included in this report?
Analysis: The image you've provided appears to be a screenshot of a document with multiple charts. However, the text is too small and blurry to read accurately. If you can provide a clearer image or more context, I might be able to help you with your question.
Extracted answer: Fail to answer
Answer format: String

---
\end{promptbox}

\newpage
\subsection{Evaluation on FinRAGBench-V}
\label{sec:app_finrag_eval}

\begin{promptbox}[unbreakable]{Prompt used for Answer Evaluation on FinRAGBench-V}
### ROLE
You are an expert evaluator. Your task is to determine if a model's generated answer is correct by comparing it to a ground truth value.

### TASK
You will be given a question, the prediction which includes reasoning steps and a final answer, and a ground_truth which is the correct answer. You must determine if the final conclusion of the prediction matches the ground_truth.

### INSTRUCTIONS
1.  **Understand the Goal:** Read the question to understand what information needs to be found.
2.  **Extract the Final Answer:** Carefully analyze the prediction. Ignore the reasoning steps and identify only the final, conclusive answer provided by the model. The answer is often at the end of the text and might be bolded.
3.  **Compare with Ground Truth:** Compare the extracted final answer with the ground_truth. Be flexible with formatting-for example, a model answer of "45 percent" should be considered a match for a ground truth of "45".
4.  **Generate Analysis:** Write a brief analysis of your finding.


### INPUTS
You will receive the data like this:
Question: [The user's question]
Ground Truth: [The expected answer]
Prediction: [The model's actual answer]

## OUTPUT FORMAT:
Your response MUST be a JSON object with two keys:
1.  score: A float, either 1.0 for a correct prediction or 0.0 for an incorrect one.
2.  reasoning: A brief, one-sentence explanation for your decision. 
\end{promptbox}

\paragraph{Calculation of page-level recall, precision, and F1.} 
To evaluate page retrieval performance, we calculate the page-level recall, precision, and F1 scores as follows. Let $\mathcal{P}_{\text{pred}}$ denote the set of predicted pages and $\mathcal{P}_{\text{gt}}$ denote the set of ground truth pages. We define true positives (TP) as $|\mathcal{P}_{\text{pred}} \cap \mathcal{P}_{\text{gt}}|$, false positives (FP) as $|\mathcal{P}_{\text{pred}} \setminus \mathcal{P}_{\text{gt}}|$, and false negatives (FN) as $|\mathcal{P}_{\text{gt}} \setminus \mathcal{P}_{\text{pred}}|$. The metrics are then calculated as:
\begin{align*}
\text{Precision} &= \frac{\text{TP}}{\text{TP} + \text{FP}}, \\
\text{Recall} &= \frac{\text{TP}}{\text{TP} + \text{FN}}, \\
\text{F1} &= \frac{2 \cdot \text{Precision} \cdot \text{Recall}}{\text{Precision} + \text{Recall}}.
\end{align*}

\paragraph{Calculation of element-level recall, precision, and F1.} 
For element-level evaluation, we assess the quality of predicted bounding boxes against ground truth annotations. Let $\mathcal{B}_{\text{gt}} = \{b_1^{\text{gt}}, b_2^{\text{gt}}, \ldots, b_n^{\text{gt}}\}$ denote the set of ground truth bounding boxes and $\mathcal{B}_{\text{pred}} = \{b_1^{\text{pred}}, b_2^{\text{pred}}, \ldots, b_m^{\text{pred}}\}$ denote the set of predicted bounding boxes, where each box is represented as $[x_1, y_1, x_2, y_2]$.

We employ Intersection over Union (IoU) as the matching criterion. For two boxes $b_i$ and $b_j$, IoU is computed as:
\begin{equation*}
\text{IoU}(b_i, b_j) = \frac{\text{Area}(b_i \cap b_j)}{\text{Area}(b_i \cup b_j)}.
\end{equation*}

A predicted box is considered a true positive if it achieves an IoU $\geq 0.5$ with a ground truth box, and each ground truth box is matched to at most one predicted box. Based on the matching results, we calculate:
\begin{align*}
\text{Precision} &= \frac{\text{TP}}{\text{TP} + \text{FP}}, \\
\text{Recall} &= \frac{\text{TP}}{\text{TP} + \text{FN}}, \\
\text{F1} &= \frac{2 \cdot \text{Precision} \cdot \text{Recall}}{\text{Precision} + \text{Recall}},
\end{align*}
where TP (true positives) is the number of successfully matched boxes, FP (false positives) is the number of unmatched predicted boxes, and FN (false negatives) is the number of unmatched ground truth boxes.

\section{Further Discussion}
\label{sec:app_further_discussion}

\subsection{Test-time Scaling of Page Navigator and Answer Sampler}
\label{sec:app_test_time_scaling}

\begin{figure}[h]
    \centering
    \includegraphics[width=\linewidth]{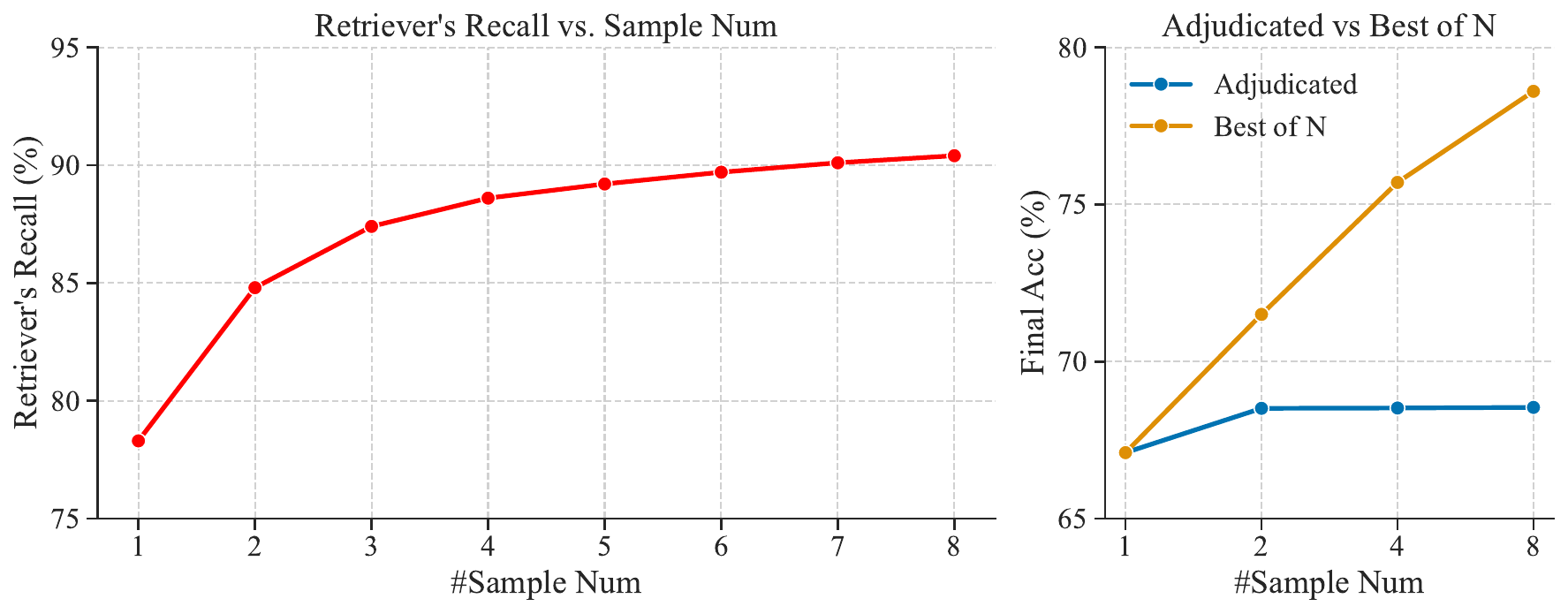}
    \caption{Test-time scaling of Page Navigator and Answer Sampler}
    \label{fig:test_time}
\end{figure}

As illustrated in Figure~\ref{fig:test_time}, we observe distinct scaling behaviors between the Page Navigator and Answer Sampler components during test-time inference.
For the Page Navigator (retrieval evidence pages), increasing the number of samples yields substantial improvements in retriever's recall performance. The recall metric rises from approximately 78\% with a single sample to over 90\% when utilizing 8 samples. However, the marginal gains diminish as the sample count increases. Notably, beyond 8 samples, the incremental improvement becomes negligible, falling below 1 percentage point. This suggests that the Page Navigator benefits significantly from test-time scaling but exhibits diminishing returns at higher sample counts.

In contrast, the Answer Sampler demonstrates markedly different scaling characteristics. When comparing the adjudicated results against the best-of-N selection, we observe that performance improvements plateau rapidly. While increasing from 1 to 2 samples produces a notable gain (approximately 5 percentage points), further scaling beyond this point yields minimal additional benefits. The adjudicated performance remains relatively flat at around 68-69\% across varying sample numbers, while even the best-of-N approach shows limited improvement beyond 2 samples. This behavior indicates that the Answer Sampler module does not require sophisticated test-time scaling strategies, and a modest sample size is sufficient to achieve near-optimal performance.
These findings suggest that computational resources for test-time scaling should be allocated primarily to the Page Navigator component, where increased sampling continues to provide meaningful improvements, rather than the Answer Sampler, where additional samples beyond 2 offer diminishing returns.

\section{Hard Cases}
\label{sec:app_hard_cases}
In this sections, we present some vision-centric cases that even our \methodname{} fail to solve. Effectively handling these cases might require either designing dedicated agentic frameworks tailored to specific visual element types or advancing the fundamental perception capabilities of backbone LLMs, which we leave for future work.

\begin{figure}[h]
\centering
\includegraphics[width=\linewidth]{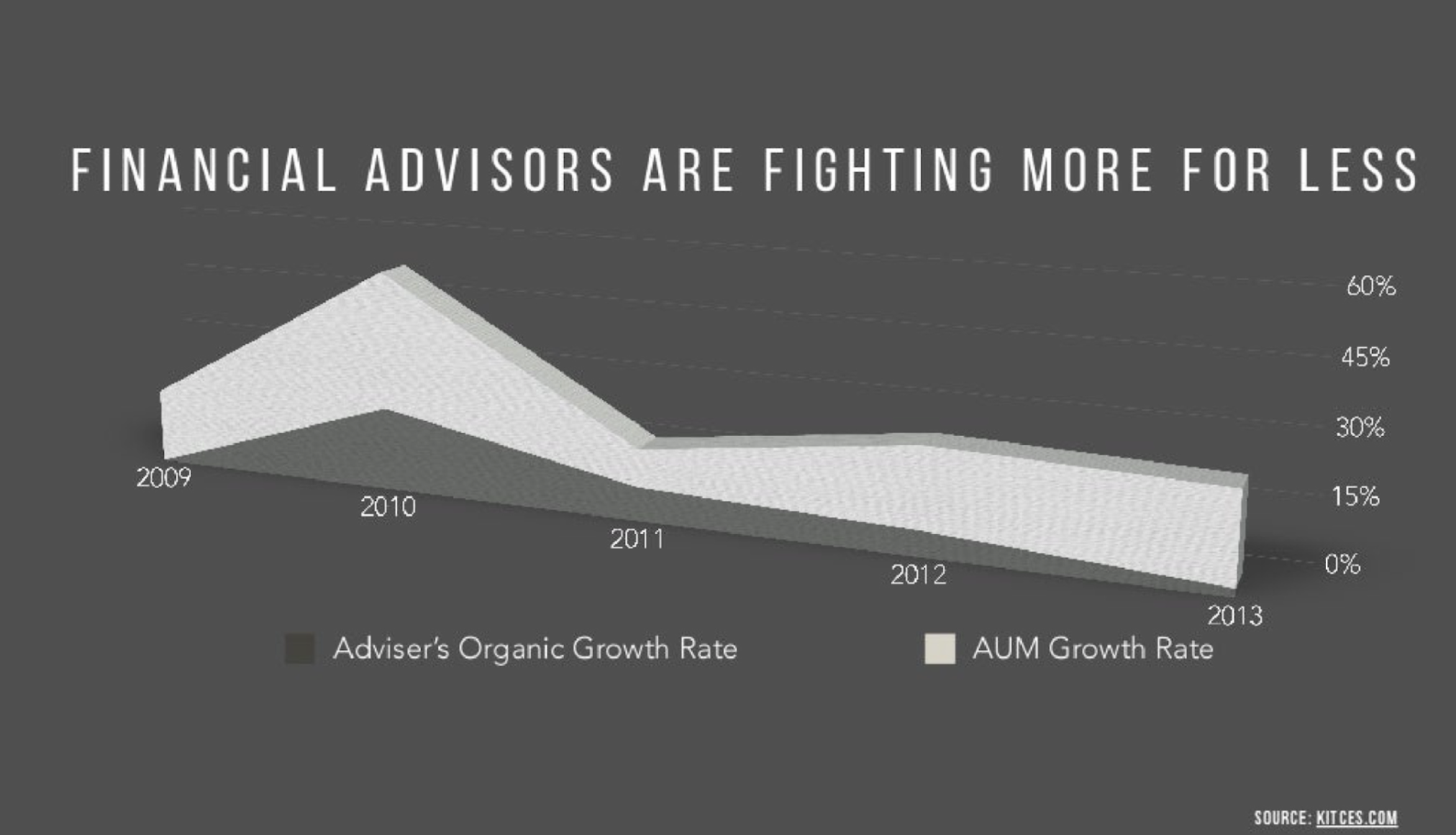}
\caption*{\textbf{Hard Case 1:} From 2009 to 2013, for the adviser's organic growth rate, how many years are higher than 2011? (\textbf{Ground Truth:} 1; \textbf{\methodname{}:} 4)}
\label{fig:hard_case_1}
\end{figure}

\begin{figure}[h]
\centering
\includegraphics[width=\linewidth]{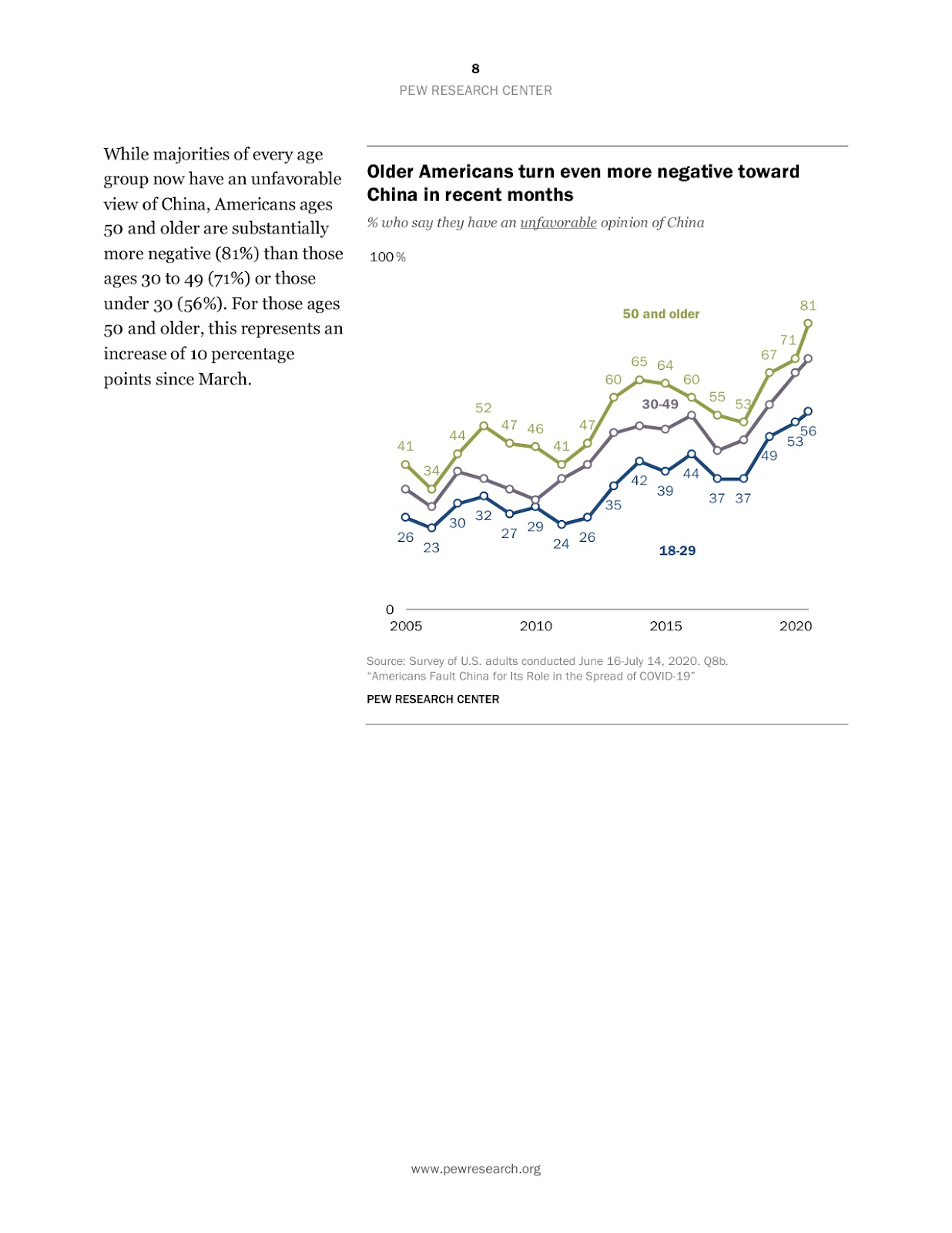}
\caption*{\textbf{Hard Case 2:} which age group experienced the greatest change in the percentage holding an "unfavorable" opinion of China between 2005 and 2010? (\textbf{Ground Truth:} 50 and older; \textbf{\methodname{}:} 30-49)}
\label{fig:hard_case_2}
\end{figure}

\begin{figure}[t]
\centering
\includegraphics[width=\linewidth]{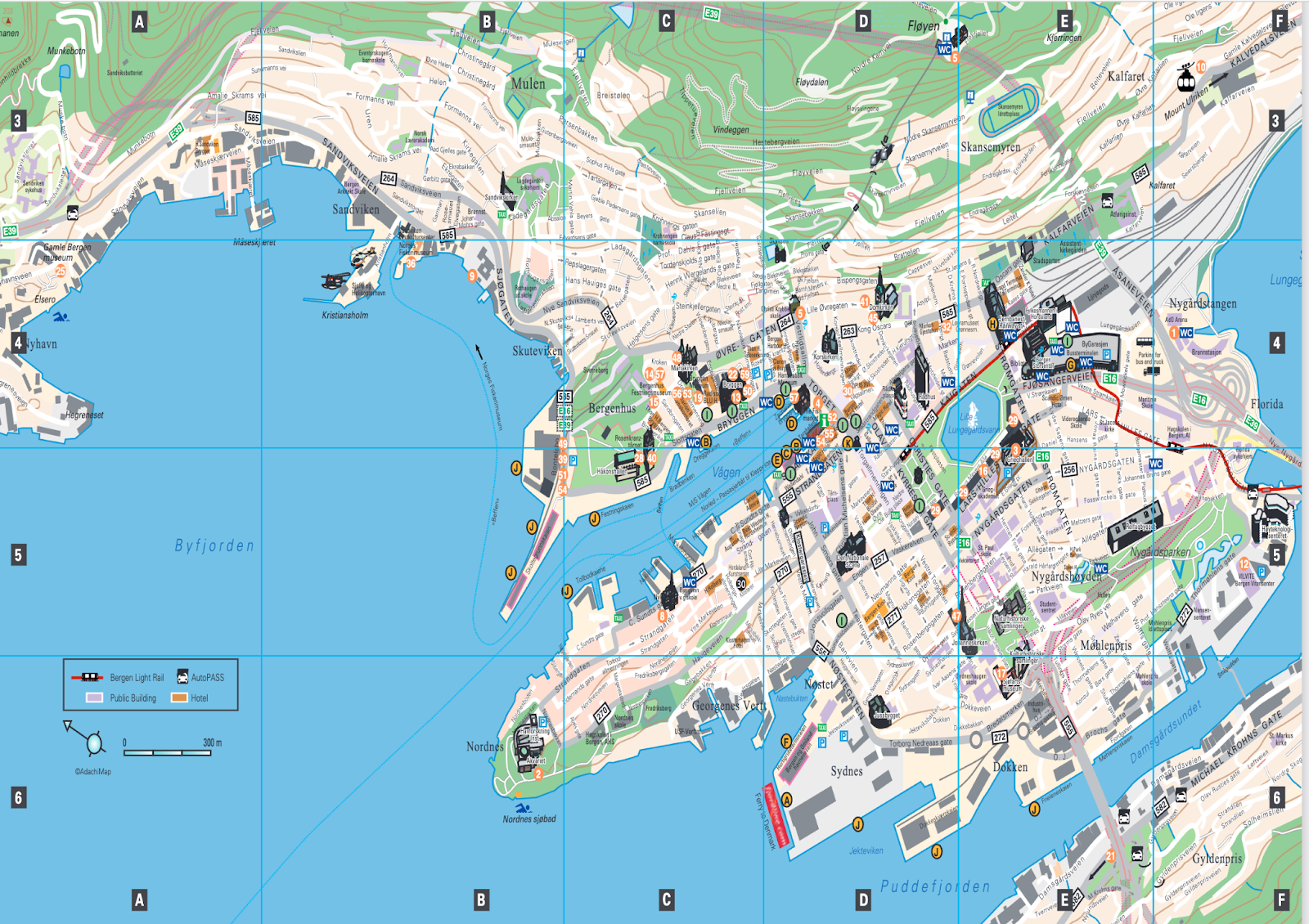}
\caption*{\textbf{Hard Case 3:} How many "WC"s are shown in the last page's map? (\textbf{Ground Truth:} 19; \textbf{\methodname{}:} 9)}
\label{fig:hard_case_3}
\end{figure}

\begin{figure}[t]
\centering
\includegraphics[width=\linewidth]{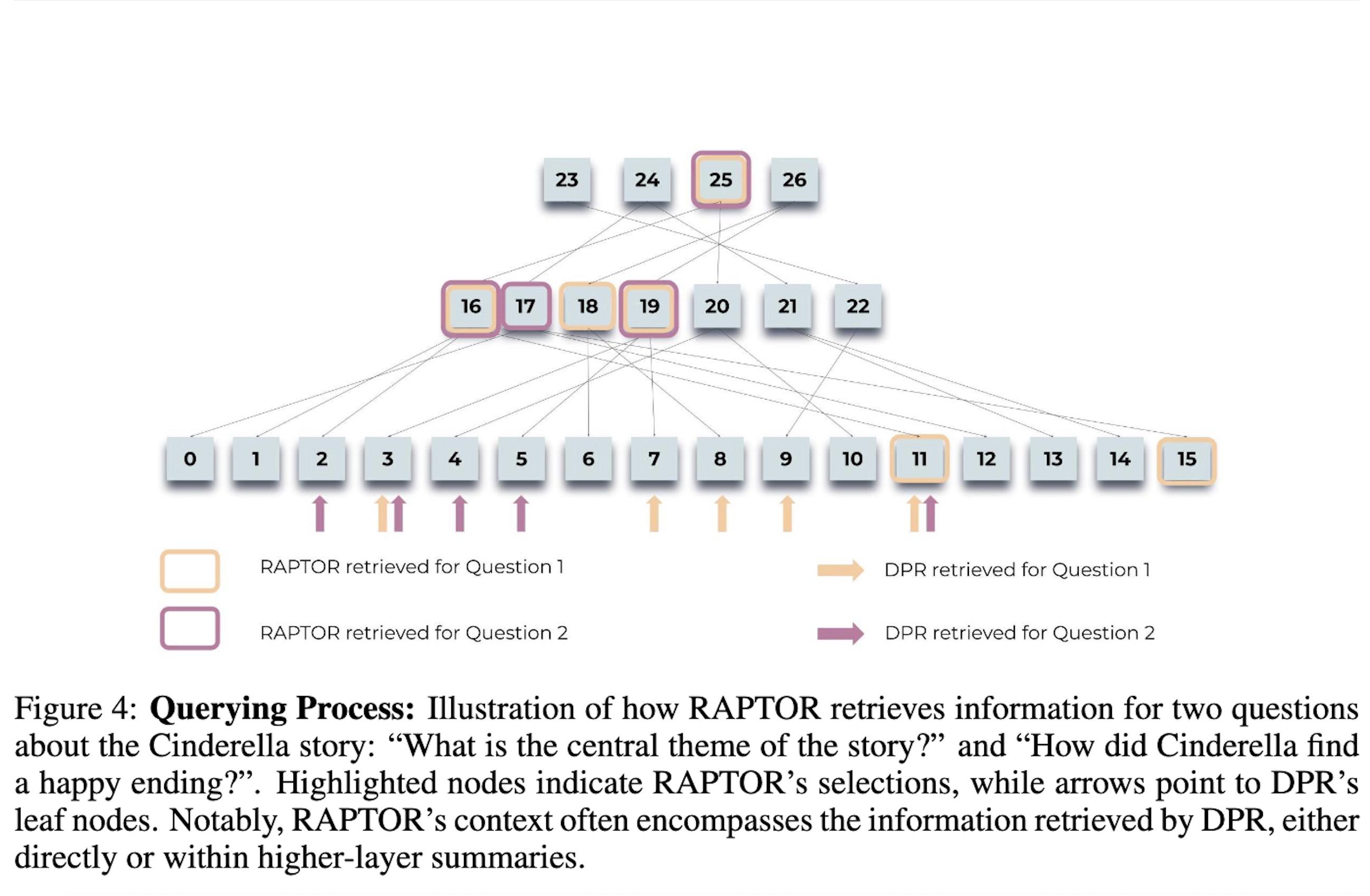}
\caption*{\textbf{Hard Case 4:} In figure 4, which nodes are retrieved by RAPTOR for both questions?
 (\textbf{Ground Truth:} [16, 19, 25]; \textbf{\methodname{}:} [1, 11, 16, 17, 20])}
\label{fig:hard_case_4}
\end{figure}

\end{document}